\pdfoutput=1
\documentclass[11pt]{article}

\usepackage[margin=1in]{geometry}
\usepackage[T1]{fontenc}
\usepackage[utf8]{inputenc}
\usepackage{natbib}
\usepackage{amsmath,amssymb,amsthm}
\usepackage{graphicx}
\usepackage{booktabs}
\usepackage{longtable}
\usepackage{multirow}
\usepackage{hhline}
\usepackage{caption}
\usepackage{subcaption}
\usepackage{enumitem}
\usepackage{placeins}
\usepackage{float}
\usepackage{xcolor}
\usepackage{xspace}
\usepackage{microtype}
\usepackage{url}
\usepackage{tikz}
\usetikzlibrary{arrows.meta,positioning,calc,fit,backgrounds}
\usepackage{hyperref}   

\hypersetup{
  colorlinks = true,
  linkcolor  = blue,
  citecolor  = blue,
  urlcolor   = blue,
  pdftitle   = {A MARL Centered Reference Architecture for Large Language
                Model Augmentation in Smart Manufacturing},
  pdfsubject = {Multiagent reinforcement learning; large language models;
                smart manufacturing},
  pdfkeywords= {large language models, multiagent reinforcement learning,
                smart manufacturing, agentic AI, foundation models,
                deployment readiness, Industry 4.0}
}

\title{A MARL Centered Reference Architecture for Large Language\\
       Model Augmentation in Smart Manufacturing}

\author{
  \textit{Fouad Bahrpeyma}\textsuperscript{*}, \textit{Dirk Reichelt} \\
  \textit{Smart Production Systems, HTW Dresden, Saxony, Germany} \\
  \texttt{Bahrpeyma@IEEE.org}, \texttt{dirk.reichelt@htw-dresden.de} \\[4pt]
  \textsuperscript{*}\,Corresponding author
}

\date{}

\begin{document}

\maketitle

\begin{abstract}
Modern manufacturing imposes six coupled demands on adaptive control: local decisions with global consequences, partial observability, nonstationarity, reflex speed response with long horizon effects, delayed and diffuse outcomes, and dynamics that resist complete explicit modeling. Multiagent reinforcement learning (MARL), especially cooperative MARL formulated as a Dec-POMDP and trained under centralized training with decentralized execution, provides a particularly natural formalism for these demands. This paper therefore adopts a MARL centered scope and asks where large language models (LLMs) should augment, interface with, train, or, in the strongest competitive case, replace the adaptive coordination core. Three analytical stages support the answer. First, a taxonomy organizes the literature through four LLM attachment points: policy, reward design, communication between agents, and hierarchical planning. Second, a conditional capability profile separates native mechanism, reported performance, formal guarantee, and engineering maturity for explicitly scoped LLM and MARL configurations. Third, an attachment decision and deployment readiness analysis identifies the conditions and present evidence associated with each role. Together these stages lead to the paper's principal contribution: a three layer MARL centered reference architecture, grounded in evidence, for semantic reasoning, adaptive cooperative control, and independently assured execution. The LLM-Augmented Dec-POMDP is a descriptive comparative notation for the same architecture. It records four attachment choices without introducing a new decision process class, solution algorithm, or theoretical guarantee. Under the reviewed evidence, conventional MARL is generally better suited to frequent, structured, decentralized coordination after task specific training, whereas LLM components are promising for semantic interpretation, reward drafting, human interaction, and slower supervisory planning. Current LLM only manufacturing controllers do not yet establish equivalence for strict real time, decentralized, safety critical control; this conclusion is bounded by the available evidence and does not assert impossibility.
\end{abstract}


\tableofcontents
\newpage

\section{Introduction}
\label{sec:intro}

A modern factory is a different object from the one for which the classical manufacturing control stack was designed \citep{KOSKY2021259,jones1998survey}. High volume, low variety production on fixed transfer lines is giving way to high mix, low volume production, with lot sizes trending toward one under mass personalization; product lifecycles are shortening; and demand is volatile enough that the production program a plant begins a week with is rarely the one it finishes the week executing \citep{shi2020smart,mittal2019smart,bicocchi2019dynamic}. The floor itself has changed along with the product mix: fixed lines are being replaced by reconfigurable cells \citep{gankin2021modular,sekar2025negotiation}, and a growing population of physically autonomous entities, automated guided vehicles and autonomous mobile robots \citep{popper2021simultaneous,malus2020real}, overhead hoist transporters \citep{ahn2021cooperative}, collaborative robots \citep{yu2021optimizing,zhang2022reinforcement}, senses, moves, and decides concurrently in a shared space. Ubiquitous instrumentation has made operational data abundant at precisely the moment when explicit process models have become hardest to keep current \citep{zuehlke2010smartfactory,wang2016implementing,wang2018knowledge,chen2017theoretical}. Disruption, machine breakdowns, rush orders, material shortfalls, has stopped being an exception to be recovered from and become a normal operating condition to be controlled through \citep{lv2024breakdown,park2019reinforcement,stricker2018reinforcement}. And the objectives themselves have multiplied: energy consumption and carbon emissions now sit alongside throughput, tardiness, and work in progress as quantities a controller is expected to manage \citep{lu2020multi,zhu2022energy,yang2025luca,luo2021dynamic}.

Stated as engineering requirements rather than as trends, this situation imposes six structural demands on any control approach, and, decisively, it imposes them jointly:

\begin{enumerate}
\item \textbf{Local decisions, global consequences.} Each machine, vehicle, or cell acts on what it can locally sense, yet the quantities that matter, throughput, makespan, work in progress, are system level properties that emerge from the interaction of all local choices \citep{gabel2009multi,dittrich2020cooperative,zhou2021multi}. No entity on the floor natively knows the global effect of its local decision.
\item \textbf{No global state.} Sensing is local, information arrives delayed, and some decisive quantities, true remaining processing time, tool wear, latent quality drift, are not observed at all \citep{oroojlooyjadid2019review,nguyen2020deep}. A control approach that presupposes an accurate global snapshot presupposes something the plant cannot supply \citep{bucsoniu2010multi}.
\item \textbf{Nonstationarity, not merely noise.} It is not that processing times have variance around a stable mean; the system itself changes, new product variants, reconfigured cells, degrading machines, other resources altering their own behavior \citep{zhang2021multi,bucsoniu2010multi,yuan2024macpro,pan2025continualsurvey}. Anything computed against a snapshot of the system begins to go stale the moment it is computed \citep{alemao2021smart}.
\item \textbf{Reflex speed response, long horizon quality.} A breakdown or a rush order demands a decision within seconds, but a good decision is one whose consequences hours downstream, the bottleneck it creates, the order it starves, have been accounted for \citep{pol2021global,luo2021real,wang2017real}. The moments that most require lookahead are the moments that least permit deliberation.
\item \textbf{Delayed, diffuse outcomes.} A dispatching choice made now surfaces as congestion later and elsewhere. Without a way of attributing a system level outcome back to the local decisions that produced it, there is no principled basis for improving those decisions \citep{foerster2018coma,sunehag2018vdn,rashid2018qmix}.
\item \textbf{Dynamics that resist explicit modeling.} Breakdown correlations, changeover interactions, human variability, and degradation are partly unknown and permanently drifting \citep{su2022deep,wang2016multi,liu2022probing}. Any approach that requires the dynamics to be written down imposes a modeling and maintenance cost that the plant pays for as long as the approach is in use \citep{sutton2018reinforcement,arulkumaran2017deep}.
\end{enumerate}

None of these demands is new, and each, taken alone, has mature candidate methods. Dispatching rules can provide very fast decisions but are commonly fixed at design time and offer limited long horizon adaptation \citep{jones1998survey,alemao2021smart}. Mathematical programming can produce high quality schedules for well specified instances, but depends on model fidelity and repeated solution after disruptions \citep{jones1998survey,liang2025marlsio}. Single agent RL learns from interaction \citep{mnih2015human,sutton2018reinforcement,arulkumaran2017deep}, yet a monolithic formulation can obscure the distributed structure and suffer from a combinatorial joint action space \citep{waschneck2018optimization,liu2020actor,bucsoniu2010multi}. Classical and model predictive control remain appropriate for many continuous, well characterized machine level dynamics \citep{KOSKY2021259,zinn2021fault}. The combined demand set therefore does not prove MARL uniquely correct; it identifies a problem structure for which cooperative MARL is a particularly natural candidate and establishes the criteria against which any alternative must be evaluated.

\begin{figure}[h!]
\centering
\begin{tikzpicture}[
  node distance=0.4cm,
  stagebox/.style={rectangle, draw, rounded corners, text width=5.4cm, minimum height=1cm, align=center, font=\small, fill=gray!5, inner sep=6pt},
  notebox/.style={rectangle, text width=5.6cm, align=left, font=\footnotesize}
]
\node (s1) [stagebox] {\textbf{Classical control} \\ fixed rules, dispatching heuristics, PID and motion control};
\node (s2) [stagebox, below=of s1] {\textbf{Optimization} \\ mathematical programming over an explicit process model};
\node (s3) [stagebox, below=of s2] {\textbf{Reinforcement learning} \\ a single agent learns a policy from interaction};
\node (s4) [stagebox, below=of s3] {\textbf{Multiagent RL (MARL)} \\ decentralized, cooperating agents formulated as a Dec-POMDP under CTDE};
\node (s5) [stagebox, below=of s4, fill=gray!15] {\textbf{LLM augmented MARL} \\ semantic reasoning, reward drafting, and human interaction layered on adaptive coordination};
\node (n1) [notebox, right=0.7cm of s1] {Fast and cheap; fixed at design time with limited long horizon adaptation};
\node (n2) [notebox, right=0.7cm of s2] {High quality for well specified instances; depends on model fidelity and re solution after each disruption};
\node (n3) [notebox, right=0.7cm of s3] {Learns from interaction; a monolithic formulation can obscure multiagent structure and suffers a combinatorial joint action space};
\node (n4) [notebox, right=0.7cm of s4] {Addresses the six manufacturing demands above jointly (Section~\ref{sec:why_marl})};
\node (n5) [notebox, right=0.7cm of s5] {Adds semantic knowledge, reward drafting, and a human interface where supported by evidence; does not yet replace task trained policies for strict real time control (Section~\ref{sec:why_llm})};
\draw[->, thick] (s1) -- (s2);
\draw[->, thick] (s2) -- (s3);
\draw[->, thick] (s3) -- (s4);
\draw[->, thick] (s4) -- (s5);
\end{tikzpicture}
\caption{Evolution of manufacturing intelligence from fixed rules to LLM augmented multiagent reinforcement learning. The progression records increasing adaptivity and the demand that motivates each stage, not a claim that later stages strictly obsolete earlier ones.}
\label{fig:evolution}
\end{figure}
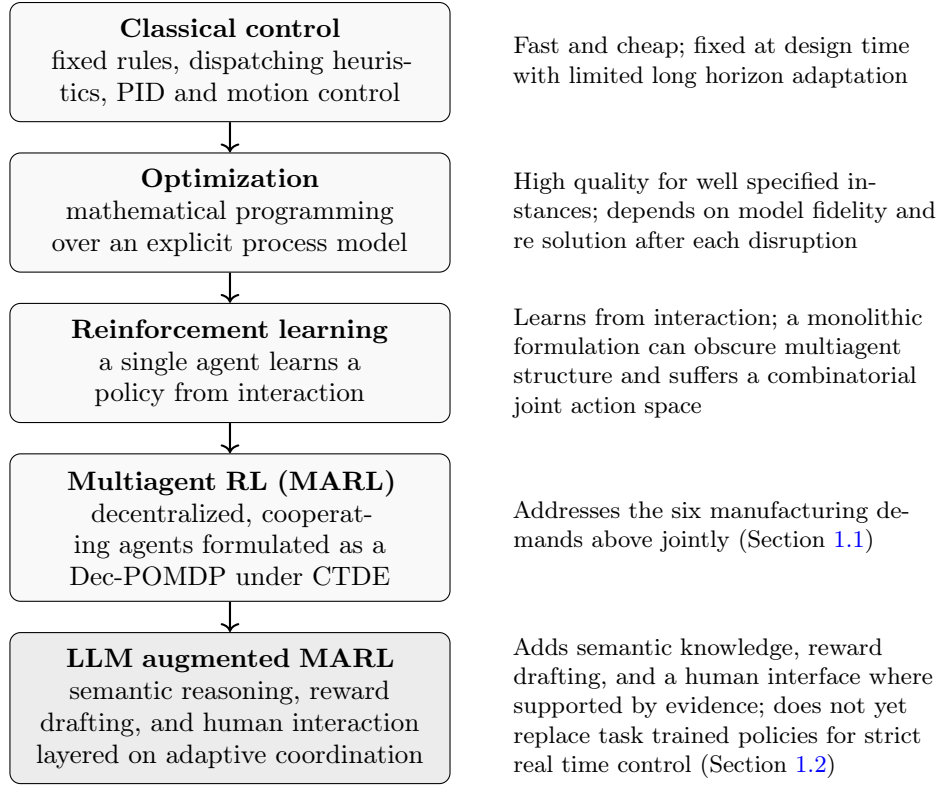

Figure~\ref{fig:evolution} summarizes this progression. It records increasing adaptivity in manufacturing control research, and the manufacturing demand that motivates each successive stage, not the obsolescence of the stages before it; each earlier method remains in industrial use for the subset of demands it was designed to meet.

\subsection{Why multiagent reinforcement learning}
\label{sec:why_marl}

MARL provides a particularly natural formalism for addressing the six demands jointly. The decentralized partially observable Markov decision process (Dec-POMDP), in which cooperative MARL is commonly posed (Section~\ref{sec:background_marl}), represents multiple decision makers with local observations, concurrent actions, stochastic dynamics, and shared or related objectives \citep{bucsoniu2010multi,zhang2021multi,oroojlooyjadid2019review}. Under CTDE, global information may be used during training while execution can rely on local observations \citep{lowe2017multi,rashid2018qmix,yu2022mappo}; learned policies amortize deliberation into training \citep{johnson2022multi,denkena2021scalable}; value factorization \citep{sunehag2018vdn,rashid2018qmix,son2019qtran,wang2021qplex} and counterfactual baselines \citep{foerster2018coma,zhang2023camarl} provide explicit credit assignment mechanisms; and model free methods learn from interaction without requiring a complete analytical transition model \citep{sutton2018reinforcement,mnih2016asynchronous}. These mechanisms align closely with the demand set, but the alignment is not itself a performance, latency, robustness, convergence, or safety guarantee. Whether a particular MARL configuration meets a factory's requirements depends on its training distribution, policy class, communications, hardware, and validation evidence \citep{gu2023safe,samak2025mixedreality,zolfagharian2024smarla}. This qualified correspondence helps explain MARL's prominence in manufacturing control research; supporting applications are reviewed in a prior review of MARL applications in smart factories \citep{bahrpeyma2022review}.

These strengths are normally realized within a numerically specified task. Conventional MARL does not, by itself, supply broad semantic knowledge, a natural language interface, faithful explanations, or automatic reward specification; these functions require additional models or engineering \citep{boggess2025explanations,kim2025talktoagent,ma2023eureka,pol2021global}. Substantial structural change may also require transfer, adaptation, or retraining \citep{yuan2024macpro,tomilin2025meal,pan2025continualsurvey,ran2024federated}. These boundaries motivate examining LLMs not as a monolithic successor to MARL, but as components that may extend a MARL centered control system.

\subsection{Why, then, large language models}
\label{sec:why_llm}

Since roughly 2023, LLMs have been used around control systems based on learning to interpret instructions \citep{huang2022languagemodels,brohan2023saycan}, plan \citep{kannan2023smartllm,prasad2023adapt,geng2025l2m2}, draft rewards \citep{ma2023eureka,xie2023text2reward,zhang2024simple}, mediate communication \citep{mandi2023roco,bae2026lmac,toquebiau2025language}, and sometimes select actions directly \citep{driess2023palme,brohan2023rt2,kim2024openvla}. These attempts are organized here relative to a MARL core \citep{sun2024llmmarl,cao2024survey}. Reward, communication, and hierarchical planning attachments primarily augment or interface with MARL \citep{zhu2025lamarl,zhuang2024yolomarl,bai2026lehca}; an LLM policy attachment is the competitive boundary case because it replaces the numeric policy $\pi_i$ \citep{yuan2025collmlight,zhang2024coela}. The latter should be evaluated under matched observations, action spaces, hardware, deadlines, training information, and safety constraints. The evidence reviewed here does not yet show that current LLM only architectures match task trained numeric policies for strict real time, decentralized, safety critical manufacturing control \citep{huang2024gammabench,pang2024illmtsc,ouyang2025nondeterminism}. Conversely, LLM components can supply semantic and human interface functions that conventional MARL does not provide by itself \citep{kim2025talktoagent,li2023tom,reflecsched2025}. The central design question is therefore normally \emph{where an LLM should attach to a MARL centered system}; full policy replacement remains an empirical alternative rather than the paper's default premise.

Much of the applied literature uses the word ``agent'' for structurally different roles, from a learned control policy \citep{lowe2017multi,rashid2018qmix} to a prompted conversational role in a language agent team \citep{wu2023autogen,li2023camel,hong2023metagpt,bandi2025agentic}. Existing surveys \citep{sun2024llmmarl,cao2024survey,li2026multirobot} catalog many of these architectures, but a manufacturing designer still needs to know whether an LLM changes the control policy, assists MARL training, enriches communication, or operates above the control loop. Section~\ref{sec:design_dimensions} supplies a scoped capability profile, and Section~\ref{sec:comparative} turns it into an attachment decision rather than a symmetric technology contest.

The question is pursued here specifically for intelligent manufacturing. The MARL algorithm and application base on which it rests was established in our earlier review of MARL applications in smart factories \citep{bahrpeyma2022review}, and is treated here as given; the present paper asks how language model components can improve, surround, train, or potentially displace parts of that base. Because the object of analysis is LLM attachment rather than MARL itself, the treatment goes deeper on attachments, their limits, and their readiness, and a comprehensive MARL survey is not repeated.

\subsection{Scope and MARL centered thesis}

The object of analysis is not an unrestricted contest between two technology labels. MARL is used as the analytical baseline because this paper focuses on decentralized adaptive manufacturing control, because the Dec-POMDP provides a natural representation of that problem \citep{oroojlooyjadid2019review,nguyen2020deep}, and because the relevant MARL literature has already been established in a prior review \citep{bahrpeyma2022review}. This choice structures the analysis but does not determine its conclusion. The Dec-POMDP describes a problem with multiple decision makers, local observations, coupled outcomes, and stochastic dynamics; it does not require the policy to be implemented by a conventional MARL network. Compact numeric networks \citep{yu2022mappo,rashid2018qmix}, language models \citep{slumbers2024leveraging,zhang2024coela}, multimodal models \citep{driess2023palme,brohan2023rt2,kim2024openvla}, hybrid policies \citep{pang2024illmtsc,wen2022mat}, and future policy classes remain admissible implementations.

Each attachment is therefore examined in turn, asking whether an LLM adds value, displaces an existing component, or should implement the policy itself. Reward design \citep{ma2023eureka,xie2023text2reward} and high level planning \citep{kannan2023smartllm,geng2025l2m2,bai2026lehca} are provisionally treated as slower complementary functions, natural language communication is conditional on bandwidth and latency \citep{mandi2023roco,bae2026lmac,toquebiau2025language}, and direct LLM policy control is evaluated as a genuine replacement option \citep{yuan2025collmlight,zhang2024coela}. The evidence required for replacement is stronger only because the policy must satisfy the coordination, timing, reproducibility, and safety requirements of the same control role, not because departure from MARL is excluded by definition. Faster local models \citep{pan2025onpremise}, constrained decoding, multimodal policies \citep{bjorck2025groot,kim2024openvla}, or MARL trained language agent teams \citep{park2025maporl,liu2026magrpo,he2025collabui} may shift these boundaries while the Dec-POMDP remains useful as a technology neutral problem formalism.

\subsection{Review scope and capability assessment method}
\label{sec:review_method}

This paper is a \emph{structured critical review and architectural synthesis}, not a preregistered systematic review or meta analysis. Its purpose is to derive and interrogate a reference architecture, rather than estimate a pooled effect size or claim exhaustive database coverage. The focal search period is January 2022 through 4 August 2026, reflecting the emergence of LLM based control and agent architectures \citep{yao2023react,brohan2023saycan,huang2022languagemodels}; earlier MARL and control works are included when needed to establish mechanisms or formal properties \citep{sunehag2018vdn,foerster2018coma,rashid2018qmix}. Candidate publications were identified iteratively from the surveys cited in Section~\ref{sec:why_llm}, backward and forward citation chaining, and keyword searches over scholarly search engines, publisher indexes, and preprint repositories. Search concepts combined variants of ``large language model,'' ``foundation model,'' or ``language agent'' with ``multiagent reinforcement learning,'' ``multiagent control,'' ``manufacturing,'' ``scheduling,'' ``robot,'' ``traffic,'' ``energy,'' ``reward design,'' ``communication,'' and ``planning.'' Because no fixed, preregistered set of bibliographic databases was queried, the paper does not claim PRISMA style completeness.

Works were included in the architecture corpus when they (i) placed an LLM at a policy, reward, communication, or hierarchical planning attachment to an RL/MARL or multiagent control problem \citep{zhu2025lamarl,zhuang2024yolomarl,bae2026lmac}; (ii) used MARL to train or assign credit within a team of language agents \citep{park2025maporl,liu2026magrpo,li2025debate}; or (iii) provided manufacturing evidence directly relevant to one of those roles \citep{gu2025llmhfs,reflecsched2025,qin2024knowledge,yang2025luca}. Studies that replace a MARL policy with an LLM or another policy class were eligible on the same basis as studies that augment MARL; they were not downgraded merely because they departed from the analytical baseline. Single agent and adjacent domain studies were retained only when they established a transferable attachment mechanism or supplied evidence for a capability not yet studied in manufacturing. Generic manufacturing chatbots, retrieval systems without a control or coordination role \citep{sharma2025rag}, and conventional MARL applications with no LLM attachment were excluded from the LLM attachment corpus, the last of these because they fall outside the object of analysis rather than because they lack merit; readers seeking that material are referred to the prior review \citep{bahrpeyma2022review}. The manufacturing readiness subset applies the narrower criterion that a work evaluates a manufacturing task or a physical multirobot architecture explicitly presented as transferable to manufacturing \citep{li2026multirobot,low2025roboticreview,krnjaic2024warehouse}; borderline cases are identified in the text rather than silently treated as factory deployments.

The capability profile in Table~\ref{tbl:capabilities} is a structured expert synthesis of this corpus. Its unit is a declared \emph{system configuration}, not the labels ``LLM'' or ``MARL'' in isolation. Practical evidence ratings use the following decision rules: S (supported) requires convergent evidence from at least two independent primary studies or benchmark families; M (mixed/conditional) denotes at least two relevant reports with inconsistent results or strong dependence on task, model, or implementation; L (limited) denotes one direct proof of concept or only indirect/adjacent domain evidence; and U denotes no eligible direct evidence located. These thresholds concern the \emph{consistency and relevance} of reported performance, not deployment realism. Engineering maturity is coded separately from the validation setting: E (emerging/conceptual), S (simulation or offline benchmark), V (physical or hardware in the loop validation), and D (similar to production or operational deployment). Formal guarantee is recorded only when a cited result establishes a property under explicit assumptions; empirical success is never upgraded to a guarantee.

The profile was produced as an authorial synthesis rather than a blinded, independently coded evidence review; consequently, no between raters reliability statistic is claimed. Representative citations in each nontrivial row make the judgments auditable, and ambiguous cells are rated toward the weaker category. Before treating the profile as a reproducible systematic instrument, two reviewers should independently recode the included studies, record disagreements, and report agreement (e.g., weighted Cohen's $\kappa$) together with a study by capability evidence ledger. In its present form, the table supports transparent architectural reasoning but not population level or statistically comparative claims.

\subsection{Principal contribution and supporting analyses}

The paper makes one principal contribution: a \emph{MARL centered reference architecture grounded in evidence for LLM augmentation in manufacturing control}. It assigns semantic reasoning, adaptive cooperative control, and assured execution to three layers with distinct timescales and assurance responsibilities, while retaining explicit alternatives for LLM policy replacement, reward design, and communication. The \emph{LLM-Augmented Dec-POMDP} is the descriptive comparative notation for this architecture: its attachment tuple $\Phi$ records which LLM roles are present and which elements of the MARL formulated control problem they affect. The structural architecture and mathematical notation are two representations of the same contribution, not independent headline claims.

Three supporting analyses establish and delimit this contribution:

\begin{enumerate}
\item \textbf{Problem and architecture foundation.} The six manufacturing demands motivate a Dec-POMDP/CTDE baseline, and the literature review identifies four recurring LLM attachments, policy \citep{zhang2024coela,yuan2025collmlight}, reward \citep{ma2023eureka,xie2023text2reward}, communication \citep{mandi2023roco,bae2026lmac}, and planning \citep{kannan2023smartllm,geng2025l2m2}, relative to that baseline (Sections~\ref{sec:intro}--\ref{sec:taxonomy}).
\item \textbf{Capability and design evidence.} A scoped capability profile distinguishes native mechanism, reported practical performance, formal guarantee, and engineering maturity, while the attachment decision framework maps application requirements to candidate configurations (Sections~\ref{sec:design_dimensions} and~\ref{sec:comparative}).
\item \textbf{Empirical boundary conditions.} The manufacturing application review, five level readiness assessment, and limitations analysis identify how much of the proposed architecture has been instantiated, where present evidence remains limited to simulation, and what observations could revise the MARL centered allocation (Sections~\ref{sec:applications}--\ref{sec:future}).
\end{enumerate}

These analyses are contributions to the argument and evidence base, but their role is cumulative: together they derive, formalize, and delimit the reference architecture as the paper's final result.

\subsection{Organization}

The paper follows the derivation summarized above. Section~\ref{sec:background} establishes the required foundations, and Section~\ref{sec:taxonomy} organizes LLM use around a MARL core. Section~\ref{sec:design_dimensions} develops the scoped evidence profile from which Section~\ref{sec:reference_architecture} derives the paper's principal contribution. Section~\ref{sec:formalism} provides comparative notation for that architecture rather than introducing a new decision process or algorithm. Section~\ref{sec:comparative} operationalizes the architecture as an attachment decision. The application, readiness, limitation, and future work sections then evaluate its current empirical support and boundary conditions.

\section{Background}
\label{sec:background}

Two bodies of work are combined in the analysis that follows, and each carries its own vocabulary. The terms used throughout the paper are fixed here. Section~\ref{sec:background_marl} states the Dec-POMDP formalism and names the three difficulties it makes explicit, partial observability, nonstationarity, and credit assignment, which are the formal counterparts of the manufacturing demands set out in Section~\ref{sec:intro}. Section~\ref{sec:background_llm} states what a large language model is and identifies the three of its properties that bear on a control loop: operation over language rather than numeric observation vectors, pretrained knowledge that may transfer without task specific training, and an inference cost that is higher than a numeric policy forward pass. Only the material required by the attachment analysis is included, and neither subsection is intended as a survey of its field.

\subsection{Multiagent reinforcement learning}
\label{sec:background_marl}

MARL problems are typically formalized as a decentralized partially observable Markov decision process (Dec-POMDP), a tuple $M = \langle N, S, \{A_i\}, T, \{R_i\}, \{O_i\}, \Omega, \gamma \rangle$, where $N$ is the number of agents; $S$ is the (generally unobserved) global state space; $A_i$ is agent $i$'s action space, with $\boldsymbol{a} = (a_1, \ldots, a_N)$ the joint action; $T(s'|s,\boldsymbol{a})$ is the state transition function; $R_i(s,\boldsymbol{a})$ is agent $i$'s (possibly shared) reward function; $\Omega$ is the observation function, generating each agent's local observation $o_i \in O_i$ from the global state; and $\gamma \in [0,1)$ is the discount factor. Each agent's policy $\pi_i(a_i|o_i)$ maps its local observation to an action (or a distribution over actions), and the joint objective in the cooperative setting on which this paper focuses is to maximize the expected discounted sum of (individual or shared) rewards across all agents.

The central difficulties this formalism makes explicit, and that recur throughout this paper, are (i) partial observability, since $O_i$ generally reveals only a fraction of $S$ to any single agent; (ii) nonstationarity, since, from any one agent's perspective, the environment's effective dynamics change as the other agents' policies change during learning; and (iii) credit assignment, since a shared reward $R_i = R$ does not, by itself, indicate which agent's action was responsible for a given outcome. These are the formal counterparts of demands (2), (3), and (5) of Section~\ref{sec:intro}, which is the sense in which the Dec-POMDP is less a modeling choice imposed on manufacturing than a transcription of the conditions a factory floor already presents. A prior review \citep{bahrpeyma2022review} provides a comprehensive treatment of the algorithms developed to address these difficulties, organized into classical coordination paradigms (independent and joint action learners), value factorization methods (VDN, QMIX, QTRAN, QPLEX), actor critic and policy gradient methods (COMA, MADDPG, MAPPO, HATRPO/HAPPO), communication methods (CommNet, TarMAC, ATOC, GraphComm), and emerging paradigms including the Multi-Agent Transformer, safe/constrained MARL, and offline MARL; the reader is referred there for details, and specific algorithms are cited below only as needed to define the LLM and MARL integration points this paper focuses on.

\subsection{Large language models and agentic AI}
\label{sec:background_llm}

A large language model is a sequence model based on a transformer and pretrained on text at internet scale, along with increasing quantities of code, structured data, and multimodal content. Its training objective is to predict the next token in a sequence. Empirical studies indicate that this process can yield broad world knowledge, the ability to learn from a few examples in context, and, with appropriate prompting or fine tuning, the ability to decompose a task described in natural language into subgoals or actions. Three properties are directly relevant to integration with MARL. First, LLMs operate over natural language, or over text serializations of structured inputs, rather than fixed numeric observation vectors. Second, pretrained knowledge may support generalization with zero or few examples to configurations not encountered during task specific training. Third, an LLM inference call is usually more computationally expensive and slower than a forward pass through the numeric policy or value networks commonly used in MARL. The first two properties can add semantic functions that lie outside the six manufacturing demands, whereas the third bears directly on the requirement for rapid response and creates a direct comparison when an LLM enters the control loop at each step.

A recent and, for this paper, important development is the emergence of \emph{agentic} LLM usage patterns, in which an LLM is not queried once for a single output but is instead embedded in a loop that interleaves reasoning, planning, tool use, and reflection on its own prior outputs \citep{bandi2025agentic}. \citep{yao2023react} introduced ReAct, in which the LLM interleaves explicit natural language ``thoughts'' with actions; \citep{shinn2023reflexion} extended this with Reflexion, in which the LLM generates a verbal self critique of a failed attempt and stores it in an episodic memory buffer to inform subsequent attempts, functioning as a language based substitute for a learned value function. These agentic patterns are the building blocks from which the multiagent LLM architectures reviewed in Section~\ref{sec:taxonomy} are constructed, and are also the building blocks of the ``agentic manufacturing'' direction discussed further in Section~\ref{sec:future}.

\section{A MARL centered taxonomy of LLM attachment architectures}
\label{sec:taxonomy}

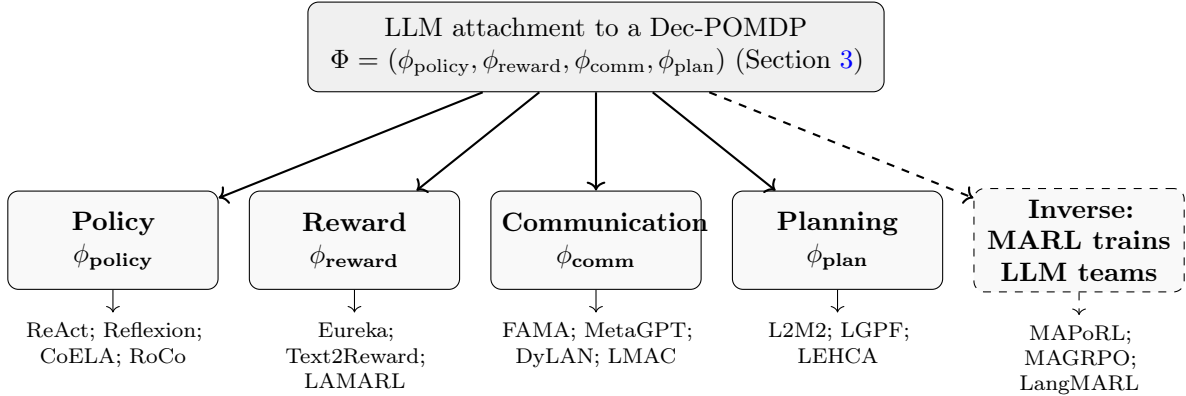
\begin{figure}[h!]
\centering
\begin{tikzpicture}[
  node distance=0.4cm,
  root/.style={rectangle, draw, rounded corners, text width=7.2cm, minimum height=1.1cm, align=center, font=\small, fill=gray!12, inner sep=6pt},
  branch/.style={rectangle, draw, rounded corners, text width=2.5cm, minimum height=1.3cm, align=center, font=\small\bfseries, fill=gray!5, inner sep=4pt},
  ibranch/.style={rectangle, draw, dashed, rounded corners, text width=2.5cm, minimum height=1.3cm, align=center, font=\small\bfseries, fill=gray!5, inner sep=4pt},
  ex/.style={rectangle, text width=2.5cm, align=center, font=\scriptsize}
]
\node (policy) [branch] {Policy \\ $\phi_{\text{policy}}$};
\node (reward) [branch, right=0.4cm of policy] {Reward \\ $\phi_{\text{reward}}$};
\node (comm)   [branch, right=0.4cm of reward] {Communication \\ $\phi_{\text{comm}}$};
\node (plan)   [branch, right=0.4cm of comm] {Planning \\ $\phi_{\text{plan}}$};
\node (team)   [ibranch, right=0.4cm of plan] {Inverse: MARL trains LLM teams};
\node (root) [root, above=1.3cm of comm] {LLM attachment to a Dec-POMDP\\ $\Phi=(\phi_{\text{policy}},\phi_{\text{reward}},\phi_{\text{comm}},\phi_{\text{plan}})$ (Section~\ref{sec:taxonomy})};
\node (policyex) [ex, below=0.3cm of policy] {ReAct; Reflexion; CoELA; RoCo};
\node (rewardex) [ex, below=0.3cm of reward] {Eureka; Text2Reward; LAMARL};
\node (commex)   [ex, below=0.3cm of comm] {FAMA; MetaGPT; DyLAN; LMAC};
\node (planex)   [ex, below=0.3cm of plan] {L2M2; LGPF; LEHCA};
\node (teamex)   [ex, below=0.3cm of team] {MAPoRL; MAGRPO; LangMARL};
\draw[->, thick] (root) -- (policy);
\draw[->, thick] (root) -- (reward);
\draw[->, thick] (root) -- (comm);
\draw[->, thick] (root) -- (plan);
\draw[->, thick, dashed] (root) -- (team);
\draw[->, thin] (policy) -- (policyex);
\draw[->, thin] (reward) -- (rewardex);
\draw[->, thin] (comm) -- (commex);
\draw[->, thin] (plan) -- (planex);
\draw[->, thin, dashed] (team) -- (teamex);
\end{tikzpicture}
\caption{A graphical taxonomy of LLM attachment points to a MARL formulated Dec-POMDP, complementing Table~\ref{tbl:formalism}. The four solid branches are attachment points on an otherwise conventional MARL system (Section~\ref{sec:formalism}); the dashed branch inverts the relationship, using MARL to train a team of LLM agents rather than augmenting a MARL policy (Section~\ref{sec:taxonomy_reverse}). Example papers are illustrative, not exhaustive; full citations appear in the corresponding subsection below.}
\label{fig:taxonomy}
\end{figure}

Following the Dec-POMDP formalism of Section~\ref{sec:background_marl}, an LLM can, in principle, be inserted at any of four points: (1) as (or within) the policy $\pi_i$ that maps observations to actions; (2) as the reward function $R_i$, i.e., generating or shaping the reward signal used to train a conventional RL policy; (3) as (or within) the observation/communication channel between agents, replacing learned numeric messages with natural language utterances; or (4) as a hierarchical layer above the Dec-POMDP, translating high level goals that are often issued by a human into subgoals or constraints for the low level agents. Section~\ref{sec:formalism} makes these four points formally precise; this section reviews the literature organized by them, extending the necessarily bounded treatment of the same material in that prior review \citep{bahrpeyma2022review}. Figure~\ref{fig:taxonomy} previews the resulting taxonomy before the subsections below examine each branch in turn.

\subsection{LLMs as the decision making policy}

The most direct integration replaces the numeric policy network $\pi_i(a_i|o_i)$ with a prompted LLM that receives a textual description of the observation and returns an action, or a natural language ``thought'' that is subsequently parsed into an action. Following \citep{sun2024llmmarl}, single agent instances of this pattern can be divided into \emph{open loop} variants, in which no environment reward updates the model (ReAct \citep{yao2023react}; Reflexion \citep{shinn2023reflexion}; ADaPT \citep{prasad2023adapt}, which recursively decomposes a task into subtasks only when a subtask proves too difficult to execute directly), and \emph{closed loop} variants, in which environment feedback is explicitly used (Refiner \citep{paul2023refiner}, in which a fine tuned LLM critiques intermediate reasoning steps; \citep{zhang2024simple}'s use of LLM generated feedback to shape intrinsic rewards for sparse reward tasks; Retroformer \citep{yao2024retroformer}, in which a frozen, large LLM acts as the policy while a smaller language model, trained via policy gradient on the environment reward, generates verbal feedback to steer it; and REX \citep{murthy2023rex}, which combines Monte-Carlo Tree Search with an upper confidence bound criterion to guide an LLM agent's search over action sequences). A related, multimodal line of work embeds language understanding directly into a trained policy network rather than relying on prompting alone: PaLM-E \citep{driess2023palme} trains an embodied multimodal LLM jointly on language and sensor/robot state tokens; SayCan \citep{brohan2023saycan} grounds an LLM's linguistic proposals in a library of learned robot skills, each with a learned affordance score, so the LLM only ever proposes actions the robot can execute; and \citep{huang2022languagemodels} show that even without fine tuning, LLMs can act as zero shot planners by decomposing natural language goals into admissible action sequences.

Extending this pattern to the multiagent case introduces the coordination and nonstationarity problems central to MARL (Section~\ref{sec:background_marl}). CoELA \citep{zhang2024coela} is a modular architecture with dedicated perception, memory, communication, planning, and execution modules, in which the LLM is embedded specifically within the memory, communication, and planning modules while perception and execution remain non LLM components. SMART-LLM \citep{kannan2023smartllm} decomposes a high level instruction into sequential phases of task decomposition, coalition formation, and task allocation across a team of robots. RoCo \citep{mandi2023roco} pairs each robotic arm with its own LLM agent, coordinating across arms, including path planning, purely through natural language dialogue. Co-NavGPT \citep{yu2023conavgpt}, in contrast, uses a single centralized LLM to assign navigation frontiers to a team of embodied agents, illustrating that the classical centralized/decentralized design axis of MARL (Section~\ref{sec:background_marl}) remains relevant even within purely LLM based architectures.

\subsection{LLMs as reward designers}

Reward engineering, and the associated credit assignment problem, is one of the most persistent bottlenecks in applying MARL to a new problem (Section~\ref{sec:background_marl}). A second architecture uses the LLM not as the policy itself but as a generator of the reward function used to train a conventional (non LLM) RL or MARL policy. Eureka \citep{ma2023eureka} combines LLM based code generation with an evolutionary search loop: the LLM proposes a population of candidate reward functions as executable code, each is used to train a policy, and the LLM iterates on the best performing candidates using both the achieved return and reward component level diagnostics as feedback. Text2Reward \citep{xie2023text2reward} is a data free method that generates dense, executable reward code directly from a natural language task description and a structured representation of the environment, reporting reward functions that match or exceed written manually expert rewards on the majority of tested manipulation tasks. Neither framework has yet been extended to the multiagent credit assignment problem specifically, an open direction revisited in Section~\ref{sec:future}.

A related but temporally distinct pattern uses the LLM only once, or only during training, rather than at every decision step. LAMARL \citep{zhu2025lamarl} uses an LLM to generate both a prior cooperative policy and a reward function before conventional MARL training begins, reporting a 185.9\% average sample efficiency improvement and, notably, both simulated and physical multirobot shape assembly experiments, one of the few LLM and MARL architectures in this review validated on hardware rather than in simulation alone (Section~\ref{sec:readiness}). YOLO-MARL \citep{zhuang2024yolomarl} follows the same one time pattern explicitly: the LLM is queried once per environment to generate strategies, interpret states, and construct planning functions, after which training proceeds with ordinary decentralized neural policies that incur no further LLM inference cost. \citep{li2025toolkit} instead keep the LLM active only during early training, using natural language strategic suggestions to mitigate the cold start problem in complex cooperative games before reverting to standard MARL exploration. Because the LLM is absent, or inactive, at deployment in all three cases, they avoid the real time latency and concerns about the cost of each decision that motivate much of Section~\ref{sec:comparative} and Section~\ref{sec:limitations}, at the cost of losing any runtime adaptivity the LLM might otherwise provide.

\subsection{LLMs as an communication between agents medium}

The classical ``learning to communicate'' branch of MARL equips agents with the ability to exchange learned, numeric messages (Section~\ref{sec:background_marl}); LLM based MARL replaces or augments this channel with natural language, with the practical benefit that a human supervisor can read the resulting protocol without an additional decoding step. Readability provides an interaction surface, but does not by itself establish that a message faithfully explains the mechanism that produced an action. FAMA \citep{slumbers2024leveraging} combines a centralized critic with natural language communication between agents, further fine tuning the LLM online so its pretrained linguistic knowledge is aligned to the functional requirements of the task environment. MetaGPT \citep{hong2023metagpt} has agents publish messages to a shared pool and selectively subscribe to messages relevant to their own role. DyLAN \citep{liu2023dylan} adjusts the communication topology between agents at inference time based on an unsupervised ``agent importance score,'' rather than fixing the topology in advance. \citep{li2023tom} incorporate Theory of Mind (ToM) modeling, in which agents' communicated messages explicitly represent beliefs about the mental states of other agents, and \citep{chen2023consensus} study how an agent's assigned ``personality'' (stubborn versus suggestible), the number of agents, and the communication network topology affect the speed and quality of consensus seeking negotiations.

A distinct role, separate from generating or interpreting runtime messages, is using the LLM to design the communication protocol itself. LMAC \citep{bae2026lmac} has an LLM reason over a natural language description of the task and each state/observation dimension to construct a communication protocol that lets agents reconstruct the underlying global state as accurately and uniformly as possible, iteratively refining that protocol against an explicit state awareness criterion. MARLIN \citep{godfrey2024marlin} instead uses dialogue mediated by an LLM for \emph{negotiation between robots}: robots negotiate natural language plans that guide a downstream MAPPO policy, with the system dynamically shifting weight from language based planning toward learned control as training progresses. \citep{toquebiau2025language} take a third, more foundational approach, training agents to produce and interpret language descriptions defined by humans of their own observations, so that language structures the agents' learned representations directly rather than being generated after the fact by a separate, frozen LLM; they report this grounding improves partner generalization and human and agent interaction relative to emergent (nonlinguistic) communication baselines. \citep{ma2026lamp} combine several of these roles in one pipeline for economic multiagent decision making: a ``Think'' stage extracts short- and long term trends from numerical observations, a ``Speak'' stage exchanges and interprets strategic natural language messages between agents, and a ``Decide'' stage fuses both into a MARL policy, with the authors explicitly comparing this hybrid against LLM only and MARL only baselines.

\subsection{Hierarchical LLM planner + MARL executor architectures}
\label{sec:taxonomy_hierarchical}

A fourth architecture keeps a conventional (non LLM) MARL system responsible for low level, real time control, while an LLM is placed above it as a high level planner. L2M2 \citep{geng2025l2m2} is a framework in which an LLM performs zero shot, high level strategic planning while a MARL system executes the resulting subgoals, reported to require less than 20\% of the training samples needed by MARL only baselines to reach comparable performance. The Language-Guided Pattern Formation framework \citep{lgpf2025swarm} applies the same general pattern to swarm robotics, translating a high level natural language description of a target swarm formation into subgoals that a MARL trained low level controller then executes as continuous motion commands. LEHCA \citep{bai2026lehca} gives this pattern its clearest value decomposition instantiation to date: an LLM ``Commander'' operating at a coarse timescale consumes structured textual summaries of the environment and generates strategic subgoals, semantic reward shaping rules, and action level constraints, which are grounded into a low level policy based on QMIX via dense auxiliary reward signals and dynamic action masking, reporting consistent gains over QMIX, QPLEX, MAVEN, and MAPPO baselines across StarCraft multiagent challenge scenarios. This architecture is treated as sufficiently important, and sufficiently distinct from the other three insertion points, to warrant its own treatment as this paper's proposed reference architecture (Section~\ref{sec:reference_architecture}), as well as the formal treatment and comparative discussion developed in Sections~\ref{sec:formalism} and~\ref{sec:comparative}.

\subsection{MARL as the training mechanism for collaborative LLM teams}
\label{sec:taxonomy_reverse}

The four patterns above all use an LLM to augment a conventional MARL system. A newer line of work inverts this relationship: it treats a team of LLM agents \emph{as} the decentralized policies of a cooperative stochastic game, and applies MARL not to augment them but to train them. MAPoRL \citep{park2025maporl} has multiple LLM agents independently answer a question, then engage in a discussion over multiple turns, and trains them after pretraining with a shared reward from a verifier that scores both answer correctness and the persuasiveness/correctness of the discussion itself, reporting that collaboration quality improves with the number of discussion turns specifically because the agents were trained jointly rather than left frozen. MAGRPO \citep{liu2026magrpo} formalizes this as a multiagent, multiple turn extension of group relative policy optimization, using centralized group relative advantages for joint credit assignment while preserving decentralized execution at inference time, evaluated on collaborative writing and coding tasks. CollabUIAgents \citep{he2025collabui} addresses multiagent credit assignment specifically for role free language agents operating across different interactive UI environments, using a credit reassignment mechanism supported by an LLM and synthesized preference data rather than environment specific rewards, and reports that the resulting system with seven billion parameters generalizes across environments better than prompting a frozen, much larger closed source model. \citep{li2025debate} apply the same underlying idea to debate specifically: a role differentiation module, learned jointly with the downstream task via MARL, counteracts the tendency of independently prompted LLM agents to converge on redundant, undifferentiated behavior. LangMARL \citep{yao2026langmarl} brings this furthest toward classical MARL machinery, introducing an explicit language space credit assignment mechanism and policy gradient style improvement operating directly on natural language trajectories rather than on numeric parameters. None of these five works targets manufacturing, and, at the time of writing, several remain preprints (Table~\ref{tbl:llmmarl}); this direction is nonetheless treated as a genuine structural addition to the taxonomy above, rather than a variant of it, since the object being optimized by MARL is the LLM policy itself rather than a separate numeric policy the LLM merely assists. It is worth noting that this line of work sits outside the rivalry and complementarity framing of Section~\ref{sec:why_llm} rather than resolving it either way: here MARL is not a candidate answer to the manufacturing demand set at all, but a training mechanism applied to a team of language agents whose task is typically reasoning or software generation rather than plant control. It is included because it is a real and rapidly growing use of the same two technologies together, and because the credit assignment machinery it borrows is exactly the machinery Section~\ref{sec:marl_capabilities} identifies as MARL's, but it is not treated as evidence bearing on which technology should control a factory.

\subsection{Common concerns: joint design, personality, and human in/on the loop operation}

Three further concerns cut across all four architectures above. First, because pretrained LLMs are typically far too large to run on board a mobile robot or embedded controller with millisecond level control loop requirements, \citep{sun2024llmmarl} propose a joint design pattern in which the LLM is used only during centralized training, e.g., as a centralized critic under the centralized training with decentralized execution (CTDE) scheme common in MARL, and its knowledge about communication and coordination is subsequently distilled into a much smaller model deployed on each agent at execution time. Second, because prompted LLM agents can be assigned a persona purely through natural language instructions, heterogeneous agent behavior, which in conventional MARL must be encoded in separate sets of learned network weights, can instead be obtained cheaply via prompt design, as demonstrated in the consensus seeking study of \citep{chen2023consensus}. Third, because the coordination channel in LLM based MARL is natural language, human operators can participate directly as a peer agent (human involvement in the loop) or as a supervisor who monitors and occasionally intervenes (human supervision of the loop) without requiring a specialized interface layer \citep{sun2024llmmarl}.

\section{MARL centered design dimensions: what an LLM attachment contributes}
\label{sec:design_dimensions}

Section~\ref{sec:taxonomy} classified the literature by \emph{where} an LLM enters a problem represented as a Dec-POMDP, as a policy, reward designer, communication medium, or hierarchical planner. This section asks what marginal capability each option adds, what it displaces, and what new constraints it introduces. MARL is the analytical baseline, not an automatically preferred implementation. Policy replacement is retained as a genuine candidate so that the analysis can support a MARL policy, a hybrid policy, partial replacement, or complete replacement according to the evidence.

The capability profile applies technology neutral operational criteria. An LLM used for reward drafting need not match a control policy's latency at each step, whereas any implementation that assumes the policy role must be compared on coordination quality, deadline compliance, scalability, reproducibility, adaptation cost, and safety under matched conditions. Table~\ref{tbl:capabilities} records the present evidence without treating complementarity, hybridization, or replacement as conclusions fixed in advance.

\subsection{Capabilities an LLM contributes}
\label{sec:llm_capabilities}

Five capabilities are identified, each already implicit in the architectures of Section~\ref{sec:taxonomy} and the background of Section~\ref{sec:background_llm}, but not previously named as a distinct axis of comparison. Four concern functions outside the demand set of Section~\ref{sec:intro}: semantic knowledge, instruction, readable communication, and reward specification. Readable language is treated as an interface capability rather than evidence of faithful explanation.

\textbf{Semantic and world knowledge.} An LLM's pretraining exposes it to a breadth of general and domain knowledge that no MARL policy, trained only on interactions within its own environment, has access to; this is what allows LLM based reward design (Eureka \citep{ma2023eureka}) and zero shot planning \citep{huang2022languagemodels} to produce plausible outputs for a configuration never encountered during training.

\textbf{Task decomposition and planning.} An LLM can decompose an ambiguous, high level goal into a structured sequence of subgoals without a manually coded task hierarchy, as in ADaPT's \citep{prasad2023adapt} as needed recursive decomposition and SMART-LLM's \citep{kannan2023smartllm} decomposition-coalition-allocation pipeline; this is the capability that the hierarchical planning attachment $\phi_{\text{plan}}$ (Section~\ref{sec:formalism}) is built around.

\textbf{Natural language communication.} An LLM can generate and interpret messages that are directly readable by humans, without a separate decoding step, as in MetaGPT's \citep{hong2023metagpt} shared message pool and FAMA's \citep{slumbers2024leveraging} negotiation protocol; this is the capability the communication attachment $\phi_{\text{comm}}$ is built around.

\textbf{Reflection and self critique.} An LLM can generate a verbal critique of its own or another agent's past decisions and use that critique to inform subsequent decisions without a gradient update, as in Reflexion's \citep{shinn2023reflexion} episodic memory of self critiques; this is a capability with no direct analogue in conventional MARL, where updating behavior in response to past outcomes requires a parameter update.

\textbf{Reward and objective specification.} An LLM can translate a natural language description of a desired behavior into an executable, dense reward function, as in Eureka \citep{ma2023eureka} and Text2Reward \citep{xie2023text2reward}; this is the capability the reward attachment $\phi_{\text{reward}}$ is built around.

\subsection{Capabilities MARL contributes}
\label{sec:marl_capabilities}

Five capabilities commonly associated with MARL are identified, each grounded in the formalism and algorithm families of Section~\ref{sec:background_marl} and the prior review \citep{bahrpeyma2022review}. These are properties of particular formulations and implementations, not invariants of every system labelled MARL. They restate, at the capability level, mechanisms through which appropriately designed MARL systems can address the six demands.

\textbf{Experiential, closed loop learning.} MARL optimizes policies through repeated interaction with an environment or logged experience. This supplies a task specific learning mechanism, but does not imply convergence, optimality, robustness outside the training distribution, or safe online adaptation unless the selected algorithm and assumptions establish those properties.

\textbf{Cooperative multiagent coordination.} Value factorization and multiagent actor critic methods provide explicit mechanisms for optimizing shared or related objectives. Some algorithms establish properties under stated assumptions, for example, QMIX's individual global maximum structural property or monotonic improvement results for particular trust region methods, but these are algorithm level results, not systemwide guarantees for arbitrary MARL deployments.

\textbf{Credit assignment.} Methods such as COMA's counterfactual baseline and QMIX's mixing network explicitly address attribution of a shared team outcome to agents' actions. Prompted critique or verbal feedback in an LLM system may supply useful heuristics, but is not presently equivalent to a generally accepted temporal and between agents credit assignment objective.

\textbf{Real time, low latency execution.} Compact trained policy networks can execute at low latency on suitable hardware. Actual end to end latency depends on network size, observation processing, communication, hardware, scheduling, and runtime software; it must be measured against the application's deadline rather than inferred from the algorithm label alone. LLM latency likewise varies sharply with model size, locality, quantization, prompt length, decoding constraints, tool calls, and serving load.

\textbf{Decentralized execution and reproducibility.} CTDE permits policies to execute from local observations after centralized training, but does not require the learned policy to be deterministic or independent of communication. A greedy action selection rule with fixed weights can be algorithmically deterministic for a fixed numeric input; stochastic policies, exploration, sensor noise, message delay, online adaptation, parallel numerical kernels, and hardware/software scheduling can still produce differences between runs. A distinction is therefore drawn between (i) \emph{policy determinism}, a property of the action selection mapping under fixed inputs and state; (ii) \emph{operational reproducibility}, an empirical property of the complete stack that spans sensing, communication, computation, and actuation; and (iii) \emph{safety guarantees}, which require independent analysis, constraints, shielding, verification, or certified execution components and do not follow from determinism alone.

\subsection{A comparative capability table}
\label{sec:capability_table}

Table~\ref{tbl:capabilities} is a structured capability profile, not a league table. The unit of comparison is a \emph{system configuration}: for LLMs, a language model component together with prompting or fine tuning, retrieval/tools, decoding, serving location, and invocation frequency; for MARL, an algorithm together with its training regime, policy class, communication assumptions, adaptation mode, and execution hardware. The baseline profiles are (a) a prompted or tuned for instructions LLM agent that may use tools but receives no environment specific policy gradient training, and (b) a task trained cooperative MARL policy under CTDE. Where a different variant materially changes an assessment, the condition is stated explicitly.

For each capability, four evidence dimensions are assessed using the selection boundaries and decision rules of Section~\ref{sec:review_method}. \emph{Native mechanism (N)} asks whether the configuration contains an explicit computational or optimization mechanism targeted at the capability: E = explicit, I = indirect or added, and A = absent in the baseline. \emph{Practical performance (P)} records S = supported, M = mixed/conditional, L = limited, or U = no eligible direct evidence located. \emph{Formal guarantee (G)} records A only when a theorem or certified property establishes the capability under explicit assumptions; N means that none is claimed at the technology family level. \emph{Engineering maturity (M)} separately records D = similar to production or operational deployment, V = physical or hardware in the loop validation, S = simulation/offline benchmark, and E = emerging/conceptual. These ordinal labels are not numerical scores and must not be aggregated into a winner. Absence of a citation or guarantee means ``not established within this review's scope,'' not impossibility.

Read against the question of Section~\ref{sec:why_llm}, the profile supports a conditional asymmetry. For frequent decisions with structured observations, a stable action interface, and task specific training data or a simulator, current evidence generally favors conventional MARL mechanisms for closed loop optimization, multiagent credit assignment, and low latency decentralized policies. For semantic interpretation, natural language interaction, task decomposition, and reward drafting, current LLM systems provide mechanisms that conventional MARL does not contain by itself. This pattern motivates, but does not prove, complementarity. It can change with the system variant: a small distilled local model with constrained decoding differs from a frontier cloud model using multistep tools, while a stochastic communicating MARL policy with online adaptation differs from a fixed greedy CTDE policy. The proposed architecture should therefore be read as a testable design hypothesis, not as a universal dominance result.

Structural reconfiguration remains an important unresolved case. A MARL policy may handle variation represented in its training distribution, while changes to entities, topology, or the state/action schema may require transfer learning, meta learning, architectural adaptation, or retraining. An LLM's pretraining and language interface may support few- or zero shot transfer, but acceptable closed loop performance and safety in reconfigured manufacturing systems have not been demonstrated by the corpus reviewed here. The profile therefore records limited evidence rather than awarding either family a categorical advantage, and Section~\ref{sec:future} proposes a matched comparison.

{\footnotesize
\setlength\LTleft{0pt}
\setlength\LTright{0pt}
\begin{longtable}{p{0.23\textwidth}p{0.18\textwidth}p{0.18\textwidth}p{0.31\textwidth}}
\caption{Conditional capability profile for two baseline configurations: a prompted/using tools LLM agent without environment specific policy gradient training and a task trained cooperative MARL policy under CTDE. Entries report native mechanism (N), practical evidence (P), formal guarantee (G), and manufacturing engineering maturity (M), using the criteria defined in the text. Ratings are scoped judgments, not aggregate scores.}
\label{tbl:capabilities} \\
\toprule
\textbf{Operational capability} & \textbf{LLM profile} & \textbf{MARL profile} & \textbf{Scope, conditions, and representative evidence} \\
\midrule
\endfirsthead
\multicolumn{4}{l}{\footnotesize\textit{Table~\ref{tbl:capabilities}, continued from previous page}} \\
\toprule
\textbf{Operational capability} & \textbf{LLM profile} & \textbf{MARL profile} & \textbf{Scope, conditions, and representative evidence} \\
\midrule
\endhead
\midrule
\multicolumn{4}{r}{\footnotesize\textit{continued on next page}} \\
\endfoot
\bottomrule
\endlastfoot
\textbf{C1.} Semantic grounding: maps language and retrieved artifacts to symbols or constraints relevant to the task & N:E; P:M; G:N; M:E & N:A; P:U; G:N; M:S & LLM evidence depends on grounding, retrieval quality, and tool interfaces; fluent output is not proof of factual grounding. MARL can receive engineered semantic features but does not create them intrinsically \citep{yao2023react,sharma2025rag}. \\
\textbf{C2.} Multistep planning and task decomposition: produces an ordered, executable subgoal structure & N:E; P:M; G:N; M:E & N:I; P:M; G:A; M:S & LLM performance is dependent on the task and prompting; hierarchical MARL supplies learned temporal abstraction when explicitly designed \citep{kannan2023smartllm,geng2025l2m2}. \\
\textbf{C3.} Human interaction and readable rationales: accepts instructions and produces messages or rationales that a person can inspect & N:E; P:S; G:N; M:E & N:I; P:L; G:N; M:S & This rating concerns interface accessibility. Faithfulness, causal fidelity, stability, and operator usefulness require separate evaluation and are not inferred from fluent text \citep{shinn2023reflexion,zhang2024coela}. \\
\textbf{C4.} Reward/objective drafting: translates a language specification into candidate executable reward code & N:E; P:M; G:N; M:E & N:A; P:U; G:N; M:S & LLM generated rewards require testing for specification gaming; conventional MARL consumes rather than normally authors the reward \citep{ma2023eureka,xie2023text2reward}. \\
\textbf{C5.} Closed loop task learning: improves a policy from environment interaction against an explicit return & N:I; P:L; G:N; M:E & N:E; P:S; G:A; M:S to D & In context memory is not equivalent to parameter learning. MARL guarantees exist only for selected algorithms and assumptions; online plant adaptation may be unsafe \citep{shinn2023reflexion,bahrpeyma2022review}. \\
\textbf{C6.} Cooperative coordination: optimizes joint behavior under coupled actions and shared/correlated returns & N:I; P:M; G:N; M:E & N:E; P:S; G:A; M:S to D & Prompted LLM teams coordinate on some benchmarks but add communication cost; cooperative MARL explicitly optimizes joint objectives \citep{zhang2024coela,slumbers2024leveraging,bahrpeyma2022review}. \\
\textbf{C7.} Temporal and interagent credit assignment: attributes delayed team return to agents and actions for learning & N:A; P:U; G:N; M:E & N:E; P:S; G:A; M:S & Verbal critique may guide revision but is not a substitute for an explicit learning objective such as counterfactual or factorized value learning \citep{bahrpeyma2022review}. \\
\textbf{C8.} Structural reconfiguration: retains acceptable performance after changes to entities, topology, or state/action schema & N:I; P:L; G:N; M:E & N:I; P:M; G:N; M:S & LLM transfer is plausible but unestablished for safety critical manufacturing; MARL can use transfer, meta learning, or retraining. Direct controlled comparisons are lacking. \\
\textbf{C9.} Deadline compliance: meets a declared end to end decision deadline at a specified load and hardware target & N:I; P:M; G:N; M:E & N:I; P:S; G:N; M:S to D & Neither label guarantees timing. Local small/distilled LLMs may meet slower loops; compact policies often suit faster loops, subject to sensing and communication overhead \citep{yuan2025collmlight}. \\
\textbf{C10.} Policy determinism: identical model state and numeric input produce the same selected action & N:I; P:M; G:N; M:E & N:I; P:S; G:N; M:S to D & Constrained/greedy decoding can reduce LLM variability; greedy MARL policies can be deterministic, while sampled policies are not. Serving and numerical kernels may still vary \citep{ouyang2025nondeterminism}. \\
\textbf{C11.} Operational reproducibility: repeated end to end trials under a declared protocol remain within tolerance & N:I; P:L; G:N; M:E & N:I; P:M; G:N; M:S to D & Must be measured across sensors, communications, compute, software, and hardware; policy determinism is neither necessary nor sufficient. \\
\textbf{C12.} Safety assurance: unsafe actions are excluded or bounded under stated assumptions & N:A; P:L; G:N; M:E & N:I; P:M; G:A; M:S to D & Neither technology is safe by default. Guarantees arise from specific constrained algorithms, shields, monitors, or certified execution layers, not from ``LLM'' or ``MARL'' generically \citep{bahrpeyma2022review}. \\
\end{longtable}
}

Figure~\ref{fig:capmatrix} distills a subset of Table~\ref{tbl:capabilities}'s rows into a single visual comparison.

\begin{figure}[h!]
\centering
\begin{tikzpicture}[
  node distance=0pt,
  hdr/.style={rectangle, draw, minimum width=2.6cm, minimum height=0.8cm, align=center, font=\small\bfseries, fill=gray!25, inner sep=3pt},
  lbl/.style={rectangle, draw, minimum width=4.2cm, minimum height=0.8cm, align=left, font=\footnotesize, fill=gray!5, inner sep=4pt, text width=4.0cm},
  cell/.style={rectangle, draw, minimum width=2.6cm, minimum height=0.8cm, align=center, font=\footnotesize, inner sep=3pt}
]
\node (h0) [lbl, fill=gray!25, font=\small\bfseries] {Operational capability};
\node (h1) [hdr, right=0pt of h0] {LLM};
\node (h2) [hdr, right=0pt of h1] {MARL};
\node (r1l) [lbl, below=0pt of h0] {Semantic knowledge (C1)};
\node (r1a) [cell, fill=gray!45, right=0pt of r1l] {M};
\node (r1b) [cell, fill=gray!10, right=0pt of r1a] {U};
\node (r2l) [lbl, below=0pt of r1l] {Planning (C2)};
\node (r2a) [cell, fill=gray!45, right=0pt of r2l] {M};
\node (r2b) [cell, fill=gray!45, right=0pt of r2a] {M};
\node (r3l) [lbl, below=0pt of r2l] {Coordination (C6)};
\node (r3a) [cell, fill=gray!45, right=0pt of r3l] {M};
\node (r3b) [cell, fill=gray!70, text=white, right=0pt of r3a] {S};
\node (r4l) [lbl, below=0pt of r3l] {Latency / deadlines (C9)};
\node (r4a) [cell, fill=gray!45, right=0pt of r4l] {M};
\node (r4b) [cell, fill=gray!70, text=white, right=0pt of r4a] {S};
\node (r5l) [lbl, below=0pt of r4l] {Credit assignment (C7)};
\node (r5a) [cell, fill=gray!10, right=0pt of r5l] {U};
\node (r5b) [cell, fill=gray!70, text=white, right=0pt of r5a] {S};
\node (r6l) [lbl, below=0pt of r5l] {Explainability (C3)};
\node (r6a) [cell, fill=gray!70, text=white, right=0pt of r6l] {S};
\node (r6b) [cell, fill=gray!25, right=0pt of r6a] {L};
\node (r7l) [lbl, below=0pt of r6l] {Deployment maturity};
\node (r7a) [cell, fill=gray!15, right=0pt of r7l] {E};
\node (r7b) [cell, fill=gray!70, text=white, right=0pt of r7a] {S to D};
\end{tikzpicture}
\caption{A capability comparison distilled from Table~\ref{tbl:capabilities} (rows C1, C2, C3, C6, C7, C9, and the engineering maturity column M aggregated across rows for the deployment row). Cell letters reproduce Table~\ref{tbl:capabilities}'s own rating codes rather than a new score: for the first six rows, S supported, M mixed/conditional, L limited, U no eligible direct evidence located (practical evidence, P); for the deployment row, E emerging/conceptual through D deployment grade (engineering maturity, M). Darker shading marks the stronger rating on each row's own scale. As in Table~\ref{tbl:capabilities}, this is a scoped, conditional profile for the two baseline configurations defined in Section~\ref{sec:capability_table}, not an aggregate score or a technology ranking.}
\label{fig:capmatrix}
\end{figure}

Appendix~\ref{app:evidence_ledger} provides study level traceability from the reviewed readiness corpus to capability rows C1 through C12 in Table~\ref{tbl:capabilities}. The ledger documents manufacturing evidence used in the synthesis; it is not a mechanical derivation of every rating, which also draws on the broader literature and algorithm specific formal results.

\section{A reference architecture for LLM and MARL systems}
\label{sec:reference_architecture}

This section presents the paper's principal contribution. The preceding analysis began from technology neutral manufacturing requirements and retained direct policy replacement as an admissible outcome. Under the evidence currently reviewed, task trained MARL has the strongest support for frequent, structured, decentralized coordination, while LLM components add semantic and human interface mechanisms and an independent execution layer supplies assurance. Layer 2 is therefore assigned to MARL as a conditional default derived from current evidence, not as a consequence of choosing MARL as the analytical baseline. The architecture is a testable structural result rather than a unique design; other policy classes should occupy Layer 2 whenever they satisfy the same operational criteria with equivalent or stronger evidence.

\subsection{Three layers, one architecture}

The reference architecture has three layers corresponding to distinct timescales and assurance responsibilities. Layer 2 addresses the six demands through task trained cooperative policies; Layer 1 adds semantic functions; and Layer 3 supplies independently engineered deadline and safety assurance that neither learned technology label provides by itself. \textbf{Layer 1 (semantic reasoning)} is realized by a selected LLM configuration and interprets natural language situations, decomposes goals, and supports operator interaction at a slower planning epoch $k$. Its outputs remain proposals subject to grounding and validation; natural language rationales are not assumed faithful. \textbf{Layer 2 (adaptive cooperative control)} is realized by a selected MARL configuration and targets coordination, credit assignment, and task specific learning conditioned on subgoal $z_k$; its latency, robustness, and reproducibility must be measured for the deployed stack. \textbf{Layer 3 (assured execution)} is realized by classical controllers, runtime monitors, action constraints, and, where required, an independently certified safety PLC. It enforces timing and safety properties under stated assumptions. Calling this layer ``assured'' rather than merely ``deterministic'' emphasizes that deterministic output alone neither establishes operational reproducibility nor guarantees safety.

\begin{figure}[h!]
\centering
\begin{tikzpicture}[
  node distance=0.55cm,
  layerbox/.style={rectangle, draw, rounded corners, text width=7.5cm, minimum height=1.3cm, align=center, font=\small, fill=gray!5, inner sep=6pt},
  sidebox/.style={rectangle, draw, rounded corners, text width=3.6cm, minimum height=0.9cm, align=center, font=\footnotesize, fill=gray!12, inner sep=4pt}
]
\node (l1) [layerbox] {\textbf{Layer 1, Semantic reasoning (LLM)} \\ interprets goal $g \in L$, aggregated state $\bar{S}_k$; outputs subgoal $z_k$ via $\phi_{\text{plan}}$};
\node (l2) [layerbox, below=of l1] {\textbf{Layer 2, Adaptive cooperative control (MARL)} \\ policies $\pi_i(a_i|o_i,z_k)$; cooperation, credit assignment, closed loop learning};
\node (l3) [layerbox, below=of l2] {\textbf{Layer 3, Assured execution (classical control)} \\ PID and motion control that meet declared deadlines; constraints, monitors, and a certified safety PLC arbitrate final actuation};
\node (rw) [sidebox, right=0.8cm of l1] {$\phi_{\text{reward}}$: offline, shapes Layer 2's training objective before deployment};
\node (cm) [sidebox, right=0.8cm of l2] {$\phi_{\text{comm}}$: NL channel within Layer 2, only if the latency budget permits};
\draw[->, thick] (l1) -- node[right, font=\scriptsize]{subgoal $z_k$} (l2);
\draw[->, thick] (l2) -- node[right, font=\scriptsize]{joint action $\boldsymbol{a}$} (l3);
\draw[->, thick, dashed] ([xshift=-1.5cm]l2.north) -- node[left, font=\scriptsize]{state summary $\bar{S}_{k+1}$} ([xshift=-1.5cm]l1.south);
\end{tikzpicture}
\caption{The proposed, testable reference architecture for LLM and MARL systems. Layer assignments reflect the scoped capability profile in Table~\ref{tbl:capabilities}, not a universal technology ranking. Reward and communication attachments are auxiliary modifications to Layer 2. A digital twin (not depicted) provides a shared training and validation substrate \citep{bahrpeyma2022review}.}
\label{fig:refarch}
\end{figure}
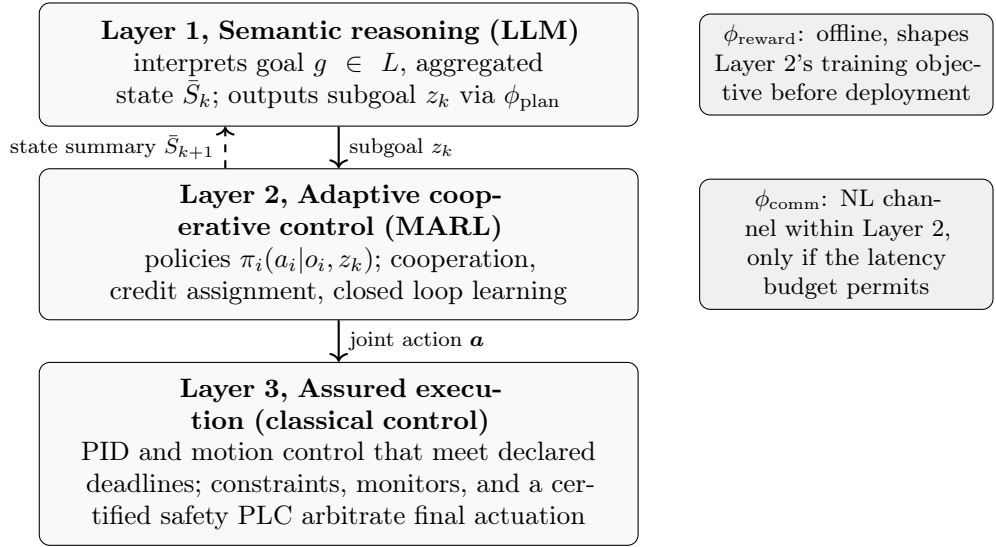

Figure~\ref{fig:refarch} depicts this architecture, together with the two auxiliary attachment points from Section~\ref{sec:formalism} that do not correspond to a distinct layer, but instead modify an existing layer's training or operation: the reward attachment $\phi_{\text{reward}}$, shown as an offline process that shapes Layer 2's training objective before deployment rather than participating in the control loop at each step, and the communication attachment $\phi_{\text{comm}}$, shown as an optional channel within Layer 2 through which agents exchange natural language, rather than learned numeric, messages. This optional channel is subject to the same decision latency criterion developed in Section~\ref{sec:comparative}: it is viable only when the decision budget of Layer 2 at each step can absorb an LLM call at every communication round; for control loops with budgets below one second that motivate placing Layer 3 beneath Layer 2 in the first place, $\phi_{\text{comm}}$ should be left null and Layer 2 should fall back to conventional, numeric MARL communication (or none), with the LLM's coordination role confined to the slower epoch $\phi_{\text{plan}}$ attachment in Layer 1. A digital twin, discussed at length in the prior review \citep{bahrpeyma2022review}, is shown as a shared substrate underlying all three layers, providing the safe training and validation environment that Layer 1's planning decisions, Layer 2's policy updates, and Layer 3's runtime monitoring checks (Section~\ref{sec:limitations}) all separately require.

\subsection{Instantiating the reference architecture}

The architecture level literature reviewed in Section~\ref{sec:taxonomy} instantiates this reference architecture only partially. L2M2 \citep{geng2025l2m2} and the Language-Guided Pattern Formation framework \citep{lgpf2025swarm} instantiate Layers 1 and 2 in full, but neither paper's evaluation setting includes a Layer 3 in the sense defined here (a certified, independently verified execution layer): L2M2 is evaluated entirely in simulation, where Layer 3 is implicitly assumed away, while \citep{lgpf2025swarm} additionally reports physical tabletop robot trials (Section~\ref{sec:readiness}) but, like L2M2, without any independently certified execution layer. GR00T N1 \citep{bjorck2025groot} instantiates a version of Layers 1 and 2 within a single embodied agent (its dual system architecture pairs a slow vision and language reasoning module with a fast diffusion transformer execution module) but is single agent and therefore has no Layer 2 in the multiagent, cooperative coordination sense used here. No existing work is known to instantiate all three layers, together with a digital twin as shared substrate, within a single manufacturing system; we consider this the most direct, actionable implication of the reference architecture for future work, and return to it in Section~\ref{sec:future}.

\section{A comparative notation for LLM attachments to a Dec-POMDP}
\label{sec:formalism}

Section~\ref{sec:reference_architecture} presented the paper's principal contribution in structural form. A comparative notation is now provided that makes the taxonomy's four attachment points explicit alongside the Dec-POMDP tuple. This notation is called the LLM-Augmented Dec-POMDP. It is a descriptive integration formalism for the reference architecture, not a new decision process class, stochastic game theory, algorithm, solution method, or independent headline contribution. Its purpose is to expose which architectural roles are active and enable precise comparison across implementations.

\subsection{Definition}

Let $L$ denote the space of natural language utterances (formally, finite sequences of tokens from the LLM's vocabulary). For comparative purposes, a Dec-POMDP with optional LLM attachments is denoted by
\begin{equation}
M^{L} = \langle N, S, \{A_i\}, T, \{R_i\}, \{O_i\}, \Omega, \gamma, L, \Phi \rangle,
\end{equation}
where $N, S, \{A_i\}, T, \{R_i\}, \{O_i\}, \Omega, \gamma$ retain their conventional meanings from Section~\ref{sec:background_marl}, and $\Phi = (\phi_{\text{policy}}, \phi_{\text{reward}}, \phi_{\text{comm}}, \phi_{\text{plan}})$ is a tuple of \emph{attachment descriptors}. The tuple $\Phi$ is architectural metadata that records where an LLM enters the system and which conventional element it affects. It does not, by itself, alter the assumptions of the underlying Dec-POMDP, establish that language interactions satisfy the Markov property, define a solution method, or transfer any convergence, optimality, or safety guarantee. Each component is either the null attachment $\bot$ or a function parameterized by an LLM of the following type:

\begin{itemize}
\item $\phi_{\text{policy}}: L \rightarrow \Delta(A_i)$, composed with a fixed verbalization function $\ell: O_i \rightarrow L$ that serializes agent $i$'s numeric observation into a natural language description, replaces the learned policy network with $\pi_i^{L}(a_i|o_i) = \phi_{\text{policy}}(\ell(o_i))$, optionally routed through an intermediate natural language ``thought'' $\xi_i \in L$ before being parsed into an action.
\item $\phi_{\text{reward}}: L \rightarrow (S \times A \rightarrow \mathbb{R})$ is a \emph{meta level} function, applied once (or infrequently, e.g., once per new task specification) prior to or during training rather than at every environment step, that maps a natural language task description $\tau \in L$ to a reward function $\hat{R}_i = \phi_{\text{reward}}(\tau)$ used in place of, or to shape, $R_i$; conventional MARL training then proceeds unmodified using $\hat{R}_i$.
\item $\phi_{\text{comm}}: O_i \rightarrow L$ generates a natural language message $m_i = \phi_{\text{comm}}(o_i)$ that is broadcast to some neighbor set $N_i \subseteq N \setminus \{i\}$, replacing (or augmenting) the observation function $\Omega$ with an augmented observation $\hat{o}_i = (o_i, \{m_j : j \in N_i\})$.
\item $\phi_{\text{plan}}: L \times \bar{S}_k \rightarrow Z$ operates at a slower planning epoch $k$ (with $k$ indexing a coarser timescale than the base decision steps of the Dec-POMDP), mapping a natural language, often issued by a human, goal $g \in L$ and an aggregated state summary $\bar{S}_k$ to a subgoal $z_k \in Z$ drawn from a task space $Z$ distinct from the low level action space; the low level policies and rewards are then conditioned on the active subgoal, $\pi_i(a_i|o_i,z_k)$ and $R_i(s,a,z_k)$, for the duration of epoch $k$.
\end{itemize}

Conventional MARL, as reviewed earlier \citep{bahrpeyma2022review}, is recovered exactly when $\Phi = (\bot,\bot,\bot,\bot)$. Each of the four subsections of Section~\ref{sec:taxonomy} corresponds to setting exactly one component of $\Phi$ to an active, parameterized by an LLM function while leaving the remaining three at $\bot$; architectures that combine more than one active component (e.g., CoELA \citep{zhang2024coela}, which uses an LLM for both decision making and communication) are naturally represented as multiple simultaneous active entries.

\subsection{Reclassifying the taxonomy}

Table~\ref{tbl:formalism} applies this notation to a representative subset of the architectures reviewed in Section~\ref{sec:taxonomy}, making explicit, for each, which component(s) of $\Phi$ are active. This reclassification has two immediate uses beyond notational tidiness. First, it makes visible a combination that this representative subset alone does not instantiate: an active $\phi_{\text{reward}}$ combined with an active $\phi_{\text{plan}}$, i.e., an LLM that both designs (or shapes) the reward function \emph{and} issues the high level subgoals the reward is defined relative to. LEHCA \citep{bai2026lehca} (Section~\ref{sec:taxonomy_hierarchical}), reviewed elsewhere in this paper but omitted from Table~\ref{tbl:formalism}'s deliberately small representative subset, is a first concrete instance: its LLM Commander issues both a subgoal and a semantic reward shaping rule at each planning epoch. What remains genuinely open, and to which this paper returns in Section~\ref{sec:future}, is the stronger form of this combination, in which the reward function is fully derived again, rather than only incrementally shaped, each time the subgoal changes. Second, it makes precise a claim that is easy to state loosely but is important to get right: L2M2 \citep{geng2025l2m2} and the Language-Guided Pattern Formation framework \citep{lgpf2025swarm} (Section~\ref{sec:taxonomy_hierarchical}) are the only architectures in Table~\ref{tbl:formalism} with an active $\phi_{\text{plan}}$ and all other components null, i.e., architectures in which the LLM's role is \emph{exclusively} hierarchical planning above an otherwise entirely conventional MARL system; every other LLM as policy or LLM as communication architecture in the table modifies the base Dec-POMDP directly, at the same timescale as the agents' own decisions, rather than through a separate, slower planning layer. This distinction is exactly the one the comparative framework of Section~\ref{sec:comparative} turns into a decision procedure.

\begin{table}[H]
\footnotesize
\centering
\caption{A representative subset of architectures classified with the LLM-Augmented Dec-POMDP comparative notation. A filled entry records an active attachment descriptor; $\bot$ records no attachment at that point. The table describes architecture and does not imply theoretical guarantees.}
\label{tbl:formalism}
\begin{tabular}{lcccc}
\toprule
\textbf{Architecture} & $\phi_{\text{policy}}$ & $\phi_{\text{reward}}$ & $\phi_{\text{comm}}$ & $\phi_{\text{plan}}$ \\
\midrule
ReAct \citep{yao2023react} & LLM & $\bot$ & $\bot$ & $\bot$ \\
Reflexion \citep{shinn2023reflexion} & LLM & $\bot$ & $\bot$ & $\bot$ \\
CoELA \citep{zhang2024coela} & LLM & $\bot$ & LLM & $\bot$ \\
RoCo \citep{mandi2023roco} & LLM & $\bot$ & LLM & $\bot$ \\
Eureka \citep{ma2023eureka} & $\bot$ & LLM & $\bot$ & $\bot$ \\
Text2Reward \citep{xie2023text2reward} & $\bot$ & LLM & $\bot$ & $\bot$ \\
FAMA \citep{slumbers2024leveraging} & $\bot$ & $\bot$ & LLM & $\bot$ \\
MetaGPT \citep{hong2023metagpt} & $\bot$ & $\bot$ & LLM & $\bot$ \\
DyLAN \citep{liu2023dylan} & $\bot$ & $\bot$ & LLM & $\bot$ \\
L2M2 \citep{geng2025l2m2} & $\bot$ & $\bot$ & $\bot$ & LLM \\
LGPF \citep{lgpf2025swarm} & $\bot$ & $\bot$ & $\bot$ & LLM \\
\bottomrule
\end{tabular}
\end{table}

\section{When and where to attach an LLM: a MARL centered decision framework}
\label{sec:comparative}

Section~\ref{sec:taxonomy} reviewed how LLMs have been combined with MARL. Given a MARL formulated manufacturing problem, this section asks whether an LLM attachment adds sufficient value to justify its cost and risk, and at which attachment point. ``No LLM'' remains a valid outcome; direct policy replacement carries a higher evidential burden than an offline reward attachment or slow supervisory planner. The decision is organized around four problem characteristics that determine the appropriate attachment and timescale.

\subsection{Four decision criteria}

\textbf{Decision latency budget.} LLM and MARL latency must be measured for the declared model, hardware, serving stack, prompt or observation length, communication load, and concurrency. Current frontier or multistep LLM systems will often exceed a budget below one second for each step, while small local, distilled, or constrained models may not. An active $\phi_{\text{reward}}$ evaluated offline or $\phi_{\text{plan}}$ evaluated at a slower epoch avoids placing language model inference on the critical path; a $\phi_{\text{policy}}$ or $\phi_{\text{comm}}$ attachment evaluated at each step requires direct worst case and tail latency validation.

\textbf{Structuredness of the state description.} MARL policies require a fixed dimensional numeric observation vector, which is straightforward to define for a well characterized problem that has already been formalized (e.g., machine occupancy or queue lengths), but becomes awkward for a state that is naturally described in free text or varies across problem instances (e.g., a maintenance log or natural language work order). LLMs, through the verbalization function $\ell$ (Section~\ref{sec:formalism}), handle the latter directly; the less structured the state description, the stronger the case for an active $\phi_{\text{policy}}$, $\phi_{\text{comm}}$, or $\phi_{\text{plan}}$ attachment.

\textbf{Availability of a well specified reward signal.} If a clear and explicitly computable performance metric already exists, such as makespan, tardiness, or energy cost, reward engineering is a comparatively minor concern and an active $\phi_{\text{reward}}$ offers limited marginal benefit over a reward specified manually. If the desired behavior is easier to describe in natural language than to specify numerically, an active $\phi_{\text{reward}}$ attachment (Eureka \citep{ma2023eureka}, Text2Reward \citep{xie2023text2reward}) may be useful.

\textbf{Need for human interaction and readable rationales.} An active $\phi_{\text{comm}}$ or $\phi_{\text{plan}}$ supplies a natural language interaction channel, while conventional MARL requires a separate interface. A readable rationale is not necessarily a faithful causal explanation. Any explanation claim should be evaluated for faithfulness, grounding, stability, operator comprehension, and override behavior.

\subsection{A decision procedure}

Table~\ref{tbl:decision} maps the four criteria onto candidate attachment patterns. A factory may need low latency coordination at each step and, simultaneously, natural language reward specification or planning intended for operators at a slower timescale. Current evidence supports $\phi_{\text{plan}}$ for infrequent semantic decisions, a task trained MARL policy as the default for frequent coordination, and an independently assured classical execution layer for final actuation. This allocation is conditional. A language model, hybrid policy, or another policy class should replace the MARL default when it demonstrates equivalent or superior control quality while meeting the same information, communication, timing, reproducibility, adaptation, and safety requirements. The framework is designed to make this common evidential standard explicit.

\begin{table}[H]
\footnotesize
\centering
\caption{A MARL centered decision framework for deciding whether and where an LLM attachment adds value.}
\label{tbl:decision}
\begin{tabular}{p{0.24\textwidth}p{0.24\textwidth}p{0.24\textwidth}}
\toprule
\textbf{Criterion} & \textbf{Favors conventional MARL ($\Phi=\bot$)} & \textbf{Favors an active LLM attachment} \\
\midrule
Decision latency budget & Decisions below one second at each step & Offline ($\phi_{\text{reward}}$) or slow epoch ($\phi_{\text{plan}}$) decisions \\
State structuredness & Fixed numeric vector, well characterized problem & Free text or variable structure state description \\
Reward specification & Clear numeric metric already exists & Desired behavior easier to state in language than in code \\
Human interaction & Separate interface acceptable & Natural language interaction or readable rationales required \\
\bottomrule
\end{tabular}
\end{table}

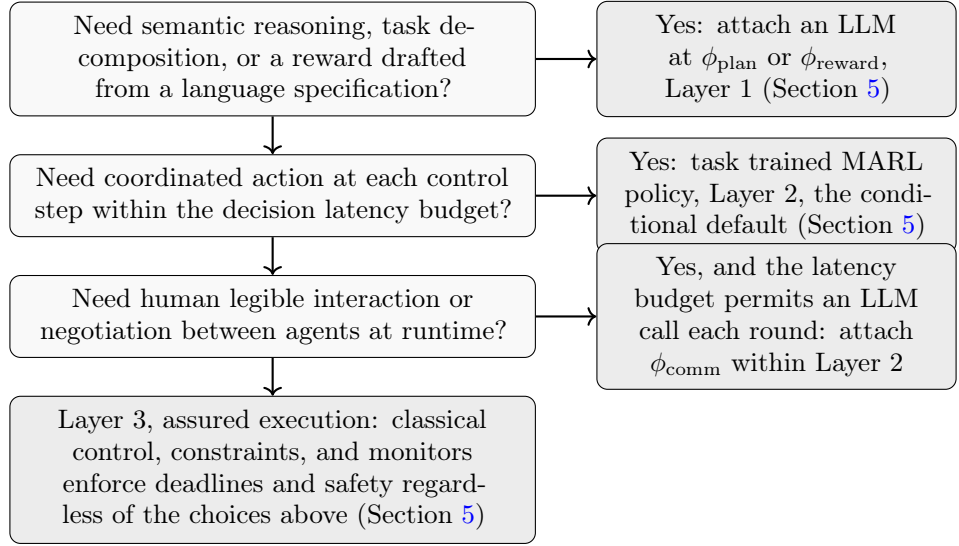
\begin{figure}[h!]
\centering
\begin{tikzpicture}[
  node distance=0.5cm,
  qbox/.style={rectangle, draw, rounded corners, text width=6.6cm, minimum height=1cm, align=center, font=\small, fill=gray!5, inner sep=5pt},
  abox/.style={rectangle, draw, rounded corners, text width=4.4cm, minimum height=1cm, align=center, font=\small, fill=gray!15, inner sep=5pt}
]
\node (q1) [qbox] {Need semantic reasoning, task decomposition, or a reward drafted from a language specification?};
\node (a1) [abox, right=0.8cm of q1] {Yes: attach an LLM at $\phi_{\text{plan}}$ or $\phi_{\text{reward}}$, Layer 1 (Section~\ref{sec:reference_architecture})};
\node (q2) [qbox, below=of q1] {Need coordinated action at each control step within the decision latency budget?};
\node (a2) [abox, right=0.8cm of q2] {Yes: task trained MARL policy, Layer 2, the conditional default (Section~\ref{sec:reference_architecture})};
\node (q3) [qbox, below=of q2] {Need human legible interaction or negotiation between agents at runtime?};
\node (a3) [abox, right=0.8cm of q3] {Yes, and the latency budget permits an LLM call each round: attach $\phi_{\text{comm}}$ within Layer 2};
\node (l3) [qbox, below=of q3, fill=gray!15] {Layer 3, assured execution: classical control, constraints, and monitors enforce deadlines and safety regardless of the choices above (Section~\ref{sec:reference_architecture})};
\draw[->, thick] (q1) -- (q2);
\draw[->, thick] (q2) -- (q3);
\draw[->, thick] (q3) -- (l3);
\draw[->, thick] (q1) -- (a1);
\draw[->, thick] (q2) -- (a2);
\draw[->, thick] (q3) -- (a3);
\end{tikzpicture}
\caption{A decision flowchart for attaching an LLM to a MARL centered manufacturing controller, operationalizing the criteria of Table~\ref{tbl:decision} and Section~\ref{sec:comparative}. Each question is evaluated independently; a single system may activate more than one attachment. Layer 3 is unconditional and independent of the Layer 1 and Layer 2 configuration chosen above it.}
\label{fig:decisionflow}
\end{figure}

Figure~\ref{fig:decisionflow} operationalizes Table~\ref{tbl:decision} as a routing procedure. The three questions are evaluated independently rather than as an exclusive branch: a single manufacturing system may need semantic planning, real time coordination, and human legible communication at once, in which case more than one attachment activates simultaneously.

\subsection{Minimum reporting checklist for normalized comparison}
\label{sec:reporting_checklist}

Comparisons among LLM, MARL, and hybrid controllers are interpretable only when the underlying configurations and operating conditions are disclosed. Table~\ref{tbl:reporting_checklist} proposes a minimum reporting checklist. It does not prescribe a preferred technology; it identifies the information needed to determine whether two results are meaningfully comparable. When conditions cannot be matched, the differences should be reported and the conclusion limited accordingly. Mean latency alone is insufficient for manufacturing control because tail latency and the frequency of missed deadlines may determine operational suitability.

\begin{table}[H]
\footnotesize
\centering
\caption{Minimum reporting checklist for normalized comparison of LLM, MARL, and hybrid manufacturing controllers.}
\label{tbl:reporting_checklist}
\begin{tabular}{p{0.20\textwidth}p{0.31\textwidth}p{0.39\textwidth}}
\toprule
\textbf{Category} & \textbf{Required fields} & \textbf{Purpose} \\
\midrule
Model configuration & Architecture, parameter count, prompting or fine tuning, distillation, quantization, numerical precision, and use of retrieval or tools & Identifies the actual system rather than attributing results to a broad technology label. \\
Execution environment & Local, edge, private cloud, or external cloud execution; processor, accelerator, memory, batch size, and concurrent load & Makes latency, cost, privacy, and scalability results interpretable. \\
Control problem & Number of agents; observation and action representations; discrete or continuous actions; partial observability; communication topology; and disruption type & Establishes whether candidate systems solve the same coordination problem. \\
Operating regime & Decision frequency, operating timescale, communication frequency, message size, prompt length, response length, tool calls, and reasoning rounds & Identifies what lies on the critical control path and how load scales with agents and communication. \\
Training and adaptation & Training data or simulator access, environment interactions, training compute, pretrained information, and adaptation or retraining budget & Normalizes differences in prior information, optimization effort, and response to reconfiguration. \\
Runtime performance & Median, 95th percentile, and 99th percentile latency; worst observed latency; deadline violation rate; task performance; and monetary or energy cost for each decision & Separates typical behavior from tail risk and relates system quality to operational cost. \\
Reliability and assurance & Evaluation seeds, repeated trials, variation between runs, failure and fallback behavior, action constraints, runtime monitors, safety layer, and certification status & Distinguishes repeatability and safety evidence from isolated benchmark success. \\
Validation context & Simulation, digital twin, hardware in the loop, physical pilot, or production; evaluation duration; and reported operational outcomes & Separates practical performance from engineering and deployment maturity. \\
\bottomrule
\end{tabular}
\end{table}

A direct comparison should match observations, admissible actions, agent count, communication budget, hardware class, deadline, safety layer, evaluation seeds, and adaptation budget whenever possible. For example, a cloud LLM invoked every thirty seconds for planning is not directly comparable with an embedded MARL policy acting every ten milliseconds. That difference concerns configuration and operating timescale rather than the technology labels alone.

\section{Applications}
\label{sec:applications}

The architectures have been reviewed (Section~\ref{sec:taxonomy}), formalized (Section~\ref{sec:formalism}), and supplied with a framework for choosing between them (Section~\ref{sec:comparative}). This section reviews where they have actually been applied, distinguishing general domain applications from the literature specific to manufacturing that is this paper's primary focus.

\subsection{General domain applications}
\label{sec:applications_general}

Reasoning and code generation benchmarks (e.g., MATH, HumanEval) have been used to evaluate multiagent LLM collaboration frameworks such as DyLAN \citep{liu2023dylan} and MetaGPT \citep{hong2023metagpt}, and dedicated multiagent game benchmarks such as $\gamma$-Bench \citep{huang2024gammabench} have been proposed specifically to quantify LLMs' strategic decision making ability in cooperative and competitive settings. \citep{guo2024embodiedllm} studied embodied LLM agents that organize themselves into effective team structures using a Criticize-Reflect protocol. Beyond embodied and reasoning benchmarks, MARL augmented with an LLM has also been applied to network wide traffic signal control, e.g., CoLLMLight \citep{yuan2025collmlight}, which uses a structured spatiotemporal graph representation to let cooperating LLM agents reason about traffic conditions at neighboring intersections, and iLLM-TSC \citep{pang2024illmtsc}, which uses an LLM to refine and improve the policy decisions of a conventional RL traffic signal controller (an instance, in the notation of Section~\ref{sec:formalism}, of an active $\phi_{\text{policy}}$ applied only as a correction after initial inference to an otherwise conventional MARL policy, rather than replacing it outright). \citep{p2penergy2025} applied a MARL framework enhanced by an LLM, guided by an expert workflow, to real time peer to peer energy trading. \citep{agentic2024supplychain} explored autonomous LLM agents capable of negotiation for consensus seeking in supply chain coordination.

\subsection{Manufacturing applications}
\label{sec:applications_manufacturing}

LLM and MARL applications specific to manufacturing remain a small but rapidly growing subset of the literature reviewed in Section~\ref{sec:applications_general}. The works below constitute the manufacturing subset located under the scope and selection procedure of Section~\ref{sec:review_method}; exhaustive database coverage is not claimed.

\textbf{Scheduling.} \citep{qin2024knowledge} proposed a MARL approach enhanced by a knowledge graph for adaptive scheduling under mass personalization, in which implicit domain knowledge extracted from historical job allocation records is used to guide each agent toward more effective scheduling policies with faster learning rates. ReflecSched \citep{reflecsched2025} addresses the dynamic flexible job shop scheduling problem in a zero shot manner: a \emph{reflection module} has the LLM analyze simulations driven by heuristics across multiple planning horizons and distill the results into a natural language ``strategic experience'' summary, while a separate \emph{decision module} conditions the LLM's final scheduling prompt on this summary, producing nonmyopic scheduling decisions without any task specific training. LUCA \citep{yang2025luca} fuses a graph neural network, an LLM providing contextual/semantic understanding of the scheduling state, and a deep RL policy network trained on a dual objective reward balancing makespan against carbon emissions, the first work known to combine an LLM and MARL architecture with a sustainability objective. \citep{gu2025llmhfs} proposed a multiagent deep RL method empowered by a pretrained LLM for hybrid flow shop scheduling, in which the LLM component achieves coordination by stage across the flow shop's sequential production stages. MARLSIO \citep{liang2025marlsio}, while not itself an LLM based method, targets the scalability limitations that make LLM based state and communication augmentation attractive in the first place, and is a natural candidate for future combination with the architectures above.

\textbf{Supply chain and energy.} The consensus seeking LLM agents of \citep{agentic2024supplychain} apply directly to inventory management coordination across warehouses and stores, and the MARL framework enhanced by an LLM for peer to peer energy trading in \citep{p2penergy2025} extends to smart grid coordination in a manufacturing context.

\textbf{Hybrid hierarchical architectures.} L2M2 \citep{geng2025l2m2} and the Language-Guided Pattern Formation framework \citep{lgpf2025swarm} (Section~\ref{sec:taxonomy_hierarchical}), while evaluated on general multiagent and swarm robotics tasks respectively rather than a named manufacturing benchmark, are directly transferable to the manufacturing scheduling and material handling problems reviewed above, since both are independent of architecture with respect to the underlying MARL executor's task.

\textbf{Foundation models and vision and language and action (VLA) models.} Beyond text only LLMs, a broader family of foundation models couples language understanding directly to visual perception and continuous motor control. RT-2 \citep{brohan2023rt2} jointly fine tunes a large vision and language model on both internet scale image and text data and robot demonstration data, treating robot actions as additional tokens in the model's vocabulary. OpenVLA \citep{kim2024openvla} is an openly available 7-billion parameter VLA model trained on nearly a million robot episodes. GR00T N1 \citep{bjorck2025groot} is a dual system VLA architecture for generalist humanoid robots that pairs a slower vision and language reasoning module with a faster diffusion transformer module for real time motor control, itself an instance of the slow reasoning/fast execution separation formalized as $\phi_{\text{plan}}$ in Section~\ref{sec:formalism}. To date, essentially all VLA research, including all three works above, targets a single embodied agent; the multiagent extension of the VLA paradigm, e.g., a foundation model that jointly perceives and coordinates the motor actions of an entire cell of collaborating manufacturing robots, remains an open problem rather than an established line of work, and is revisited in Section~\ref{sec:future}.

\textbf{Agentic manufacturing.} The broader ``agentic AI'' literature (Section~\ref{sec:background_llm}) proposes a substantially richer notion of an agent than the Dec-POMDP policy $\pi_i$ of Section~\ref{sec:background_marl}: one that reasons, plans, retrieves relevant information, communicates, executes, learns, and reflects on its own past decisions \citep{bandi2025agentic}. Read this way, several architectures already reviewed in Section~\ref{sec:taxonomy} are early, if not always explicitly labeled, instances of agentic manufacturing: ReAct's \citep{yao2023react} interleaved thoughts correspond to reasoning; SMART-LLM's \citep{kannan2023smartllm} decomposition-coalition-allocation pipeline corresponds to planning; CoELA's \citep{zhang2024coela} memory module corresponds to retrieval; RoCo's \citep{mandi2023roco} dialogue between robotic arms corresponds to communication; and Reflexion's \citep{shinn2023reflexion} verbal self critique corresponds to reflection. What is new is not any single one of these components, individually reviewed in Section~\ref{sec:taxonomy}, but the framing of a manufacturing agent as a cognitive architecture built from all of them jointly, rather than as a single learned function $\pi_i$. This reframing raises a question that remains unresolved: how the formal multiagent guarantees of conventional MARL (e.g., QMIX's individual global maximum property, or the monotonic joint policy improvement guarantee of HATRPO/HAPPO and the Multi-Agent Transformer, all reviewed in the prior review \citep{bahrpeyma2022review}) transfer, if at all, to a team of agentic, LLM based reasoners whose ``policy'' is an emergent property of a loop for reasoning, planning, retrieval, communication, execution, learning, and reflection rather than a single differentiable function.

Table~\ref{tbl:llmmarl} summarizes the architectures and applications reviewed in Sections~\ref{sec:taxonomy} and~\ref{sec:applications}.

{\footnotesize
\begin{longtable}{ p{0.03\textwidth} p{0.14\textwidth}  p{0.14\textwidth}  p{0.12\textwidth}  p{0.13\textwidth}  p{0.05\textwidth}  p{0.22\textwidth} }
\caption{Overview of LLM and MARL integration architectures and applications reviewed in this paper.}
\label{tbl:llmmarl} \\
\hline
\textbf{Cat.} & \textbf{Reference} & \textbf{LLM Role} & \textbf{RL/MARL Backbone} & \textbf{Domain} & \textbf{Train.} & \textbf{Contribution} \\
\hline
\hline
\endfirsthead
\multicolumn{7}{l}{\footnotesize\textit{Table~\ref{tbl:llmmarl}, continued from previous page}} \\
\hline
\textbf{Cat.} & \textbf{Reference} & \textbf{LLM Role} & \textbf{RL/MARL Backbone} & \textbf{Domain} & \textbf{Train.} & \textbf{Contribution} \\
\hline
\hline
\endhead
\hline
\multicolumn{7}{r}{\footnotesize\textit{continued on next page}} \\
\endfoot
\hline
\endlastfoot
\multirow{3}{*}{\rotatebox{90}{\scriptsize Open loop policy}}
 & \citep{yao2023react} & Reasoning/Decision & None (prompting) & General reasoning & No & Interleaves thought and action generation \\ \nopagebreak \cline{2-7}
 & \citep{shinn2023reflexion} & Decision + self critique & None (prompting) & General decision & No & Verbal RL via episodic memory of reflections \\ \nopagebreak \cline{2-7}
 & \citep{prasad2023adapt} & Task decomposition & None (prompting) & General planning & No & As needed recursive task decomposition \\ \hline
\multirow{4}{*}{\rotatebox{90}{\scriptsize Closed loop policy}}
 & \citep{paul2023refiner} & Feedback/critique & Fine tuned LLM & General reasoning & Yes & Reasoning feedback on intermediate steps \\ \nopagebreak \cline{2-7}
 & \citep{zhang2024simple} & Reward/credit assignment & Sparse reward RL & General RL & Partial & shaped by an LLM intrinsic reward \\ \nopagebreak \cline{2-7}
 & \citep{yao2024retroformer} & Policy (frozen) + critic LM & Policy gradient & General decision & Yes (LM) & Retrospective verbal feedback trains a smaller critic LM \\ \nopagebreak \cline{2-7}
 & \citep{murthy2023rex} & Policy search & MCTS + UCB & General decision & No & Tree search over proposed by an LLM actions \\ \hline
\multirow{6}{*}{\rotatebox{90}{\scriptsize Embodied / VLA}}
 & \citep{driess2023palme} & Policy (multimodal) & End to end trained & Robot control & Yes & Joint language + sensor token training \\ \nopagebreak \cline{2-7}
 & \citep{brohan2023saycan} & Planner (grounded) & Learned skill library & Robot manipulation & Yes & LLM proposals grounded by affordance scores \\ \nopagebreak \cline{2-7}
 & \citep{huang2022languagemodels} & Zero shot planner & None & Embodied tasks & No & Decomposes NL goals into admissible actions \\ \nopagebreak \cline{2-7}
 & \citep{brohan2023rt2} & Policy (VLA) & End to end trained & Robot control & Yes & Actions as tokens in VLM vocabulary \\ \nopagebreak \cline{2-7}
 & \citep{kim2024openvla} & Policy (VLA) & End to end trained & Robot manipulation & Yes & Open 7B-parameter VLA baseline \\ \nopagebreak \cline{2-7}
 & \citep{bjorck2025groot} & Policy (dual system VLA) & Diffusion transformer & Humanoid manipulation & Yes & Slow reasoning + fast diffusion execution \\ \hline
\multirow{4}{*}{\rotatebox{90}{\scriptsize Multiagent policy}}
 & \citep{zhang2024coela} & Decision, Comm., Memory & Modular (LLM + non LLM) & Embodied household MAS & Yes & Modular cooperative embodied agent (CoELA) \\ \nopagebreak \cline{2-7}
 & \citep{kannan2023smartllm} & Decision, Planning & None & Multirobot task planning & No & Decomposition/coalition/ allocation phases \\ \nopagebreak \cline{2-7}
 & \citep{mandi2023roco} & Decision, Planning & None & Multirobot arm collaboration & No & Dialectic negotiation between arm agents \\ \nopagebreak \cline{2-7}
 & \citep{yu2023conavgpt} & Planning (centralized) & None & Multirobot navigation & No & Single LLM assigns frontiers to team \\ \hline
\multirow{2}{*}{\rotatebox{90}{\scriptsize Reward}}
 & \citep{ma2023eureka} & Reward generation & Evolutionary search + RL & Manipulation, locomotion & Yes (RL) & coded by an LLM reward with evolutionary refinement \\ \nopagebreak \cline{2-7}
 & \citep{xie2023text2reward} & Reward generation & RL & Manipulation, locomotion & Yes (RL) & Data free dense reward code from NL goal \\ \hline
\multirow{3}{*}{\rotatebox{90}{\scriptsize During training}}
 & \citep{zhu2025lamarl} & Policy prior + reward & MARL & Multirobot shape assembly & Yes (RL) & One time prior policy/reward gen.; sim.\ + physical robots \\ \nopagebreak \cline{2-7}
 & \citep{zhuang2024yolomarl} & Strategy/state/plan gen. (once) & MARL & Cooperative games & Yes (RL) & Single LLM query before training; no runtime LLM cost \\ \nopagebreak \cline{2-7}
 & \citep{li2025toolkit} & Exploration guidance & MARL & Cooperative games & Yes (RL) & NL suggestions mitigate cold start exploration \\ \hline
\multirow{9}{*}{\rotatebox{90}{\scriptsize Communication}}
 & \citep{slumbers2024leveraging} & Decision, Communication & Centralized critic & Text games, driving & Yes & NL communication with online LLM alignment \\ \nopagebreak \cline{2-7}
 & \citep{hong2023metagpt} & Code gen., Communication & None & Software engineering & No & Shared message pool with subscription \\ \nopagebreak \cline{2-7}
 & \citep{liu2023dylan} & Decision, Communication & None & Reasoning, coding & No & Dynamic communication topology, agent importance score \\ \nopagebreak \cline{2-7}
 & \citep{li2023tom} & Decision, Comm., ToM & None & Path planning & No & Belief modeling of other agents \\ \nopagebreak \cline{2-7}
 & \citep{chen2023consensus} & Decision (negotiation) & None & Consensus seeking & No & Personality/topology effects on negotiation \\ \nopagebreak \cline{2-7}
 & \citep{bae2026lmac} & Protocol design & MARL (comm.\ module) & MARL benchmarks (StarCraft II) & Yes (RL) & LLM designs communication protocol for state recovery \\ \nopagebreak \cline{2-7}
 & \citep{godfrey2024marlin} & Negotiation (planning) & MAPPO & Multirobot navigation & Yes (RL) & LLM negotiation weight anneals toward learned control \\ \nopagebreak \cline{2-7}
 & \citep{toquebiau2025language} & Communication (grounding) & MARL & Embodied coordination & Yes (RL) & Agents learn to produce/interpret NL observation descriptions \\ \nopagebreak \cline{2-7}
 & \citep{ma2026lamp} & Reasoning, Comm., Decision & MARL & Economic decision making & Yes (RL) & Think-Speak-Decide pipeline; compared vs.\ LLM only/MARL only \\ \hline
\multirow{3}{*}{\rotatebox{90}{\scriptsize Hierarchical}}
 & \citep{geng2025l2m2} & High level planning & MARL (low level) & General multiagent tasks & Yes (RL) & LLM plans, MARL executes; sample efficient \\ \nopagebreak \cline{2-7}
 & \citep{lgpf2025swarm} & High level planning & MARL (low level) & Swarm robotics & Yes (RL) & NL pattern description mapped to swarm subgoals \\ \nopagebreak \cline{2-7}
 & \citep{bai2026lehca} & Commander (subgoals, shaping) & QMIX (low level) & StarCraft multiagent challenge & Yes (RL) & Semantic reward shaping + dynamic action masking \\ \hline
\multirow{5}{*}{\rotatebox{90}{\scriptsize MARL trains LLM teams}}
 & \citep{park2025maporl} & Policy (trained by MARL) & RL training after initial deployment & Math/NLI reasoning & Yes (RL) & Multiple turn discussion trained jointly via verifier reward \\ \nopagebreak \cline{2-7}
 & \citep{liu2026magrpo} & Policy (trained by MARL) & Group relative PO & Writing, coding & Yes (RL) & Multiagent multiple turn GRPO extension \\ \nopagebreak \cline{2-7}
 & \citep{he2025collabui} & Policy (trained by MARL) & RL + credit reassignment & Across environments UI agents & Yes (RL) & Process level credit reassignment, role free agents \\ \nopagebreak \cline{2-7}
 & \citep{li2025debate} & Policy (trained by MARL) & RL (role embeddings) & Collaborative debate & Yes (RL) & Learned role differentiation prevents behavioral collapse \\ \nopagebreak \cline{2-7}
 & \citep{yao2026langmarl} & Policy (trained by MARL) & Language space policy grad. & General coordination & Yes (RL) & Agent level language credit assignment \\ \hline
\multirow{6}{*}{\rotatebox{90}{\scriptsize Outside manufacturing}}
 & \citep{huang2024gammabench} & Evaluation/benchmark & N/A & Multiagent games & No & Benchmarks LLM strategic decision making \\ \nopagebreak \cline{2-7}
 & \citep{guo2024embodiedllm} & Decision, Communication & None & Embodied teams & No & Criticize-Reflect self organization \\ \nopagebreak \cline{2-7}
 & \citep{yuan2025collmlight} & Decision (cooperative) & Fine tuned lightweight LLM & Traffic signal control & Yes & Spatiotemporal graph, adaptive reasoning depth \\ \nopagebreak \cline{2-7}
 & \citep{pang2024illmtsc} & Policy improvement & RL (baseline) & Traffic signal control & Yes (RL) & LLM refines RL policy decisions \\ \nopagebreak \cline{2-7}
 & \citep{p2penergy2025} & Decision guidance & MARL + expert workflow & P2P energy trading & Yes & guided by an LLM expert workflow for MARL trading agents \\ \nopagebreak \cline{2-7}
 & \citep{agentic2024supplychain} & Negotiation/consensus & None & Supply chain coordination & No & Consensus seeking framework for SCM \\ \hline
\multirow{5}{*}{\rotatebox{90}{\scriptsize Smart factory sched.}}
 & \citep{qin2024knowledge} & State/knowledge augmentation & MARL & Adaptive job scheduling & Yes (RL) & Reward shaping guided by a knowledge graph \\ \nopagebreak \cline{2-7}
 & \citep{reflecsched2025} & Decision (reflection) & None (zero shot) & Dynamic FJSP & No & Reflection module distills across horizons strategic experience \\ \nopagebreak \cline{2-7}
 & \citep{yang2025luca} & Semantic state augmentation & GNN + DRL & Carbon aware FJSP & Yes (RL) & Fuses GNN structure, LLM semantics, and a DRL policy \\ \nopagebreak \cline{2-7}
 & \citep{gu2025llmhfs} & By stage coordination & MADRL & Hybrid flow shop scheduling & Yes (RL) & Pretrained LLM coordinates production stages \\ \nopagebreak \cline{2-7}
 & \citep{liang2025marlsio} & ,  (non LLM baseline) & MAPPO & Large scale FJSP & Yes (RL) & Structural information decomposition for scalability \\ \hline
\end{longtable}
}

\section{Deployment readiness: a retrospective assessment}
\label{sec:readiness}

Section~\ref{sec:applications_manufacturing} reviewed which architectures that augment MARL with an LLM have been applied to manufacturing problems; this section asks how close that literature is to actual industrial use, a question the architecture level review above cannot answer on its own. Readiness level scales are established practice in other engineering disciplines (e.g., NASA/DARPA Technology Readiness Levels), and an analogous notion has begun to appear for reinforcement learning in adjacent domains such as UAV control, but no such scale appears to have been proposed, and applied retrospectively to an existing corpus, specifically for MARL augmented with an LLM in manufacturing. We propose one here.

\subsection{A five level scale}

Five levels are defined, ordered by increasing evidence of real world validation:

\begin{itemize}
\item \textbf{Level 1, Simulation only.} The system is trained and evaluated entirely in a simulated environment (a discrete event simulator, a game engine, or a manually coded numeric environment), with no digital twin fidelity claims and no physical hardware involved.
\item \textbf{Level 2, Validated in a digital twin.} The system is trained and/or evaluated in a high fidelity digital twin of a specific physical system, with an explicitly reported or estimated simulation to reality gap, but has not yet been executed on physical hardware.
\item \textbf{Level 3, Physical pilot.} The system has been executed on physical hardware (a real robot, a real production cell) in a limited, noncontinuous trial, but has not been integrated into a live, ongoing production process.
\item \textbf{Level 4, Production deployment.} The system runs continuously, or on an ongoing operational basis, as part of a live manufacturing process, with reported operational outcomes (e.g., throughput, downtime) rather than only benchmark metrics.
\item \textbf{Level 5, Certified safety critical operation.} The system has undergone, or is explicitly designed to undergo, a formal certification or regulatory approval process (analogous to aviation's EASA Level 1/2 machine learning guidance, discussed in the prior review \citep{bahrpeyma2022review}) prior to deployment in a safety critical role.

\end{itemize}

\subsection{Retrospective scoring}

Table~\ref{tbl:readiness} scores every application in the included manufacturing subset according to the strongest validation setting reported by each paper. Eight of the nine included works are at Level 1. The Language-Guided Pattern Formation framework \citep{lgpf2025swarm} reports a Level 3 physical pilot on tabletop swarm robots, although its task is not a manufacturing benchmark. LAMARL \citep{zhu2025lamarl}, which is outside the manufacturing subset, also reports physical multirobot experiments and illustrates that LLM assistance is not inherently confined to simulation.

The prior MARL review contains a Level 4 production example \citep{bedorf2024fab}. This case is an existence proof that conventional MARL has crossed that threshold in at least one reported setting; it is not a controlled rate comparison with the smaller LLM augmented corpus. The two corpora differ in age, size, application composition, publication incentives, access to hardware and proprietary production data, and opportunity for industrial validation. Unpublished industrial systems may also be missing. The observed difference is therefore a snapshot of the published evidence located under this review's scope, not proof that systems augmented with LLMs are intrinsically less deployable.

\begin{table}[H]
\footnotesize
\centering
\caption{Retrospective readiness snapshot for the included manufacturing corpus. Scores record reported validation settings and do not estimate intrinsic deployability.}
\label{tbl:readiness}
\begin{tabular}{lcl}
\toprule
\textbf{Reference} & \textbf{Level} & \textbf{Basis} \\
\midrule
\citep{qin2024knowledge} & 1 & Simulated job allocation scheduling only \\
\citep{reflecsched2025} & 1 & Discrete event simulation only \\
\citep{yang2025luca} & 1 & Simulated flexible job shop benchmark only \\
\citep{gu2025llmhfs} & 1 & Simulated hybrid flow shop benchmark only \\
\citep{liang2025marlsio} & 1 & Simulated large scale FJSP benchmark only \\
\citep{agentic2024supplychain} & 1 & Generated/simulated negotiation data only \\
\citep{p2penergy2025} & 1 & Simulated energy trading environment only \\
\citep{geng2025l2m2} & 1 & General multiagent simulation benchmarks only \\
\citep{lgpf2025swarm} & 3 & Policies transferred to physical ``maru'' tabletop swarm robots \\
\bottomrule
\end{tabular}
\end{table}

Table~\ref{tbl:readiness} should be interpreted as a transparent and updateable snapshot of the included evidence. It records what each publication reports, not the latent potential of its technology family. A future physical pilot or production report would update the distribution directly, but the present sample is too small and heterogeneous to support statistical comparison of deployment rates.

The study level basis for each readiness assignment, its corresponding capability evidence, and the reported or missing metrics are recorded in Appendix~\ref{app:evidence_ledger}.

\section{Limitations}
\label{sec:limitations}

The predominantly Level 1 scores motivate examination of practical barriers, but do not establish that those barriers are intrinsic to every LLM configuration or absent from every MARL system. Their effect depends on model size, deployment location, hardware, operating timescale, communication pattern, and assurance architecture. The principal barriers reported or implied by the included studies are discussed.

\textit{Real time constraints.} As formalized in Section~\ref{sec:comparative}, most manufacturing scheduling and transportation tasks require decisions on the order of milliseconds to seconds, whereas LLM inference, even for a single forward pass, is typically an order of magnitude slower. Eight of the nine manufacturing applications in Table~\ref{tbl:readiness} are evaluated in discrete event simulation rather than under hard real time constraints, and even the ninth exception, which was validated on physical hardware, \citep{lgpf2025swarm} is not itself a manufacturing benchmark; whether the joint design/distillation approach discussed in Section~\ref{sec:taxonomy} (or the fine tuned, lightweight model used by CoLLMLight \citep{yuan2025collmlight} for real time traffic control) can bring latency from LLM inference in the control loop down to the range below one second required for a Level 3 or Level 4 deployment (Section~\ref{sec:readiness}) remains an open empirical question for an actual manufacturing setting.

\textit{Continuous action spaces.} Many manufacturing control problems (e.g., AMR routing bids, OHT rebalancing, reviewed in the prior review \citep{bahrpeyma2022review}) require continuous valued actions, whereas LLMs natively generate discrete text tokens. Existing workarounds, reformulating continuous control as a multiple choice problem, or replacing an LLM's final layers with a regression head that must then be trained in the target environment, either lose precision or partially forfeit the zero-/few shot advantage that motivates using an LLM in the first place (Section~\ref{sec:comparative}).

\textit{Reliability, hallucination, and safety verification.} An incorrect scheduling or routing decision on a real production line has direct financial and physical safety consequences. None of the works specific to manufacturing in Table~\ref{tbl:readiness} report any formal safety verification of the LLM's output before execution, an omission that stands in contrast to the safe/constrained MARL literature (MACPO, MAPPO-Lagrangian) surveyed in the prior review \citep{bahrpeyma2022review}. Action filtering/masking, already standard practice in several non LLM MARL works, is a natural and low cost safeguard that any deployment augmented with an LLM should retain.

\textit{Cost.} Unlike a numeric policy network, whose inference cost is a fixed, small number of matrix multiplications, an LLM call incurs a compute or API cost for each token that scales with the length of both the prompt and the generated response. \citep{pan2025onpremise} model the cost tradeoff between commercial LLM API subscriptions and on premises deployment of open source models, finding that on premises deployment only breaks even at a scale of usage that a single production line's control loop, running continuously, would reach quickly, but that a smaller or pilot deployment might not; agentic architectures (Section~\ref{sec:applications_manufacturing}), which reason over multiple tool use or reflection steps per decision rather than a single forward pass, compound this cost further. None of the works in Table~\ref{tbl:readiness} report the monetary or energy cost of using their LLM component for each decision.

\textit{Privacy.} Sending a factory's live production schedule, machine state, or process parameters to an external LLM API discloses commercially sensitive, potentially trade secret, operational data to an external party. \citep{pan2025onpremise}'s on premises deployment analysis, discussed above for its cost implications, is motivated in large part by exactly this data sovereignty concern, and is directly complementary to the federated MARL literature surveyed in the prior review \citep{bahrpeyma2022review}, which was motivated by the same unwillingness to share raw production data across organizational boundaries; no work has yet been found that combines federated or on premises LLM deployment specifically with the manufacturing scheduling architectures of Section~\ref{sec:applications_manufacturing}.

\textit{Knowledge freshness.} An LLM's parametric knowledge is fixed at training time and does not automatically reflect a specific factory's current machine population, product mix, or process parameters. \citep{sharma2025rag} surveys retrieval augmented generation (RAG) as the standard mitigation for this staleness problem; none of the works in Table~\ref{tbl:readiness}, however, incorporate a RAG component to ground the LLM's reasoning in the specific, current state of the factory it is meant to control.

\textit{Policy determinism, operational reproducibility, and safety.} These properties must not be conflated. A greedy numeric MARL policy can be algorithmically deterministic for fixed weights, model state, input, and numerical environment, whereas stochastic policies and exploration are not. Repeated LLM calls can vary because of sampling, batching, and floating point execution even at nominal temperature zero; constrained decoding or a controlled local serving stack can reduce but does not automatically eliminate system level variation. \citep{ouyang2025nondeterminism} report substantial variation in repeated code generation queries. Operational reproducibility must therefore be evaluated end to end, including sensing, communications, serving, online adaptation, and hardware/software scheduling, for both system families. Safety is a separate property: neither repeatable behavior nor a deterministic policy is necessarily safe, and assurance must come from verified constraints, shielding, runtime monitors, or certified execution components. These distinctions should be part of any Level 5 certification argument.

\textit{Prompt attacks.} Because the LLM based communication and human interface architectures of Section~\ref{sec:taxonomy} accept natural language as an input, whether from another agent, a human operator, or, in principle, any system able to inject text into that channel, they inherit the prompt injection vulnerability class documented extensively in the wider LLM security literature. \citep{gulyamov2026promptinjection} review this vulnerability class in detail, including indirect injection, in which malicious instructions are embedded in content the LLM processes (e.g., a sensor log or a work order description) rather than typed directly by an attacker, concluding that no single defensive layer reliably prevents all such attacks. This is a materially different threat model from the adversarial perturbation literature familiar from conventional deep learning, and has not yet been considered by any of the works specific to manufacturing in Table~\ref{tbl:readiness}.

\textit{Reward hacking and specification gaming.} LLM generated reward functions (Eureka \citep{ma2023eureka}, Text2Reward \citep{xie2023text2reward}, Section~\ref{sec:taxonomy}) are optimized by an evolutionary or gradient based search loop that can exploit unintended loopholes in the generated reward code, a risk compounded in the multiagent case by the credit assignment problem (Section~\ref{sec:background_marl}). Verifying an LLM generated multiagent reward function against the intended global objective before deployment is an open problem.

\textit{Benchmark and dataset scarcity.} The great majority of the LLM and MARL literature surveyed in Section~\ref{sec:applications_general} evaluates on embodied household tasks, text/reasoning benchmarks, or traffic simulators; only the nine works in Table~\ref{tbl:readiness} evaluate with a simulator specific to manufacturing, and none on a physical shop floor. The manufacturing community lacks an equivalent of the game- and household task benchmarks that have driven progress in general purpose LLM and MARL.

\textit{More agents is not automatically better.} A separate risk, specific to the multiagent LLM orchestration patterns of Sections~\ref{sec:taxonomy} and~\ref{sec:taxonomy_reverse}, is assuming that adding agents, communication rounds, or joint training based on MARL necessarily improves on a single strong LLM. \citep{tian2025orchestration} compare multiple turn multiagent LLM orchestration against single model inference across several benchmarks and find the advantage is dependent on the task: orchestration helps when a task decomposes naturally into specialized subtasks, but adds coordination overhead without corresponding benefit otherwise. This is a direct caution for the comparative decision framework of Section~\ref{sec:comparative}: a hybrid or multiagent architecture should be justified by the structure of the problem, not assumed superior by default.

\textit{Comparative notation and readiness evidence limitations.} Two supporting instruments delimit the principal architectural contribution. The LLM-Augmented Dec-POMDP is descriptive comparative notation for \emph{where} an LLM attaches. It does not alter the underlying stochastic game theory or provide a solution algorithm, performance guarantee, or convergence guarantee. The readiness scale is based on a small corpus of nine scored works and has not been validated against practitioner judgments. It therefore supplies provisional evidence about the architecture's current instantiation gap rather than an independently validated measurement instrument.

\section{Future directions}
\label{sec:future}

Five directions are highlighted that follow directly from the formal and empirical contributions above, rather than repeating the broader future directions discussion (digital twins, continual learning, agentic manufacturing, foundation models) already provided in our earlier review \citep{bahrpeyma2022review}.

\textit{Falsification and future replacement of the MARL default.} The proposed allocation would be weakened or falsified for a given application if an LLM policy, hybrid policy, or another policy class achieved equivalent or superior coordination while satisfying the same operational requirements. A decisive experiment should follow Table~\ref{tbl:reporting_checklist}: candidates should use matched observations, admissible actions, agent count, communication budget, hardware class, deadline, safety layer, evaluation seeds, and adaptation budget. Reports should disclose parameter count, execution location, inference hardware, operating timescale, communication frequency, training effort, median latency, tail latency, deadline violations, and variation between runs. Evaluation before and after structural reconfiguration would test whether the adaptation advantage sometimes attributed to language models offsets the cost of placing them in the policy role. A positive result for an alternative policy would narrow or overturn the Layer 2 MARL assignment without invalidating the Dec-POMDP as a description of the underlying problem.

\textit{Complete reward derivation under $\phi_{\text{plan}}$.} LEHCA \citep{bai2026lehca} combines reward design and planning attachments through incremental reward shaping. A stronger design would derive the reward function anew when the active subgoal changes. This would allow the reward structure, rather than only an additive shaping term, to track a change from throughput maximization to energy minimization without requiring manual specification for every transition. This stronger combination remains open.

\textit{Formal guarantees for systems with an LLM attachment.} None of the convergence or optimality guarantees available for conventional MARL currently extends generally to a system with an active $\Phi$. Given the nondeterminism discussed in Section~\ref{sec:limitations}, even the appropriate definition of a guarantee remains an open question.

\textit{Manufacturing systems grounded through retrieval augmented generation.} None of the works in Table~\ref{tbl:readiness} uses retrieval augmented generation to ground LLM reasoning in the current state of a specific factory. Combining a policy, communication, or planning attachment with retrieval from live machine status, work orders, and maintenance logs is a direct extension of the reviewed architectures.

\textit{A benchmark for LLM and MARL in manufacturing.} Without a shared manufacturing benchmark analogous to those used for games and household tasks, successive studies cannot be compared under common conditions and progress through the proposed readiness levels is difficult to track.

\section{Conclusion}
\label{sec:conclusion}

This paper uses MARL as an analytical baseline for adaptive manufacturing coordination because the six demands motivate the Dec-POMDP and because a prior review supplies the relevant MARL evidence base \citep{bahrpeyma2022review}. The baseline organizes the analysis but does not determine the implementation selected for the final architecture. The Dec-POMDP describes the control problem, whereas MARL, language models, hybrid policies, and future policy classes are candidate solution methods subject to common operational criteria.

The taxonomy, capability profile, attachment decision framework, reporting checklist, and readiness assessment are successive parts of one derivation. They indicate that task trained cooperative MARL is generally better supported for frequent, structured, decentralized coordination in the reviewed evidence, whereas LLM components are promising for slower semantic interpretation, task decomposition, natural language interaction, readable rationales, and reward drafting. This is a conditional allocation, not a universal technology ranking. Natural language output is treated as an interaction surface; faithful explanation requires separate evidence.

The paper's principal contribution is the resulting three layer MARL centered reference architecture: a selected LLM configuration for semantic reasoning, a selected MARL configuration for adaptive cooperative control, and an independently assured execution layer for deadlines and safety. The LLM-Augmented Dec-POMDP is the mathematical representation of this same architecture, with $\Phi$ encoding policy, reward, communication, and planning attachments; it is not proposed as a separate learning algorithm. Policy determinism, operational reproducibility, explanation faithfulness, and safety assurance remain distinct responsibilities. The readiness assessment shows that most works in the included manufacturing corpus report simulation level validation and that no reviewed work tests the complete three layer system. This is a snapshot of published evidence, not proof of intrinsic deployability. Validation of the architecture requires normalized comparisons using the reporting checklist, explicit safety cases, reconfiguration tests, and physical manufacturing evaluation.

\appendix
\section{Study level evidence ledger}
\label{app:evidence_ledger}

Table~\ref{tbl:evidence_ledger} links the nine studies scored in Table~\ref{tbl:readiness} to the capability identifiers in Table~\ref{tbl:capabilities}. ``NR'' means that the item was not reported in the publication material examined. A capability identifier records that the study supplied relevant practical evidence; it does not mean that the study alone determines the corresponding profile rating. The corpus contains six direct manufacturing or supply chain applications, one adjacent energy application, two transferable general multiagent or swarm benchmarks, and two informative comparators that do not contain an LLM attachment. These scope distinctions are retained rather than treating every row as equivalent evidence.

{\tiny
\setlength\LTleft{0pt}
\setlength\LTright{0pt}
\setlength{\tabcolsep}{2pt}
\begin{longtable}{|p{0.13\textwidth}|p{0.16\textwidth}|p{0.15\textwidth}|p{0.035\textwidth}|p{0.25\textwidth}|p{0.205\textwidth}|}
\caption{Evidence ledger for the studies included in the retrospective readiness corpus. Attachment labels follow Section~\ref{sec:formalism}; capability identifiers refer to Table~\ref{tbl:capabilities}; readiness levels refer to Table~\ref{tbl:readiness}.}
\label{tbl:evidence_ledger} \\
\hline
\textbf{Study and scope} & \textbf{Attachment type or comparator} & \textbf{Capability evidence considered} & \textbf{Rea\-di\-ness} & \textbf{Key reported metrics or outcomes} & \textbf{Important unreported items} \\
\hline
\endfirsthead
\multicolumn{6}{l}{\tiny\textit{Table~\ref{tbl:evidence_ledger}, continued from previous page}} \\
\hline
\textbf{Study and scope} & \textbf{Attachment type or comparator} & \textbf{Capability evidence considered} & \textbf{Rea\-di\-ness} & \textbf{Key reported metrics or outcomes} & \textbf{Important unreported items} \\
\hline
\endhead
\hline
\multicolumn{6}{r}{\tiny\textit{continued on next page}} \\
\endfoot
\hline
\endlastfoot
\citep{qin2024knowledge}; manufacturing scheduling & No LLM attachment; knowledge graph state and reward augmentation of MARL & C5 learning; C6 coordination; C8 adaptation & 1 & Episodes to convergence, learned return, makespan and comparison with individual RL and five heuristic rules across dynamic simulation cases & LLM model and serving fields not applicable; end to end latency, tail latency, physical hardware, safety assurance and production outcomes NR \\ \hline
\citep{reflecsched2025}; manufacturing scheduling & Policy and planning attachment through hierarchical reflection and final LLM decisions & C1 semantic use; C2 planning; C8 adaptation & 1 & Makespan based relative percentage deviation and win rate; 71.35\% win rate and 2.755\% relative percentage deviation reduction against direct LLM baselines & Parameter count, execution location, inference hardware, median and tail latency, safety layer and physical validation NR \\ \hline
\citep{yang2025luca}; manufacturing scheduling & Semantic state augmentation feeding a graph RL policy; closest to $\phi_{\mathrm{policy}}$ & C1 semantic use; C5 learning; C8 adaptation & 1 & Makespan and carbon emissions on synthetic and public data; average 4.1\% and up to 12.2\% lower makespan than the strongest reported comparator on synthetic data at the same emission level & Median and tail latency, deadline violations, between run variation, safety layer and physical validation NR \\ \hline
\citep{gu2025llmhfs}; manufacturing scheduling & $\phi_{\mathrm{policy}}$ through LLM enhanced state representation and action selection; conventional event driven agent communication & C1 semantic use; C5 learning; C6 coordination; C8 adaptation & 1 & Makespan across 330 simulated instances; improvement above 8\% in most instances against rules, genetic programming, DRL and MAPPO comparators; generalization tests & Parameter count, serving location, median and tail latency, deadline violations, safety layer and physical deployment NR \\ \hline
\citep{liang2025marlsio}; manufacturing scheduling & No LLM attachment; MAPPO structural information comparator & C5 learning; C6 coordination; C8 scalability & 1 & Makespan or objective quality, computation efficiency and generalization across synthetic and public large scale flexible job shop instances & LLM fields not applicable; physical validation, operational latency distribution, safety assurance and production outcomes NR \\ \hline
\citep{agentic2024supplychain}; supply chain case study & $\phi_{\mathrm{policy}}$ and $\phi_{\mathrm{comm}}$ through autonomous negotiation and consensus & C2 planning; C3 human readable interaction; C6 coordination & 1 & Inventory performance and bullwhip effect; negotiation frameworks reduced the bullwhip effect relative to restocking and centralized demand comparators & Parameter count, local or cloud execution, latency distribution, communication cost, safety layer and physical operational deployment NR \\ \hline
\citep{p2penergy2025}; adjacent energy application & Training time policy guidance from LLM generated expert strategies, followed by MARL imitation & C1 semantic strategy generation; C5 learning; C6 coordination; C8 adaptation; C12 constraint relevant evidence & 1 & Economic cost, voltage violation rate, convergence and stability in simulated test systems; lower cost and violation rate than reported baselines & Model size, serving location, median and tail decision latency, physical grid trial and certified safety assurance NR \\ \hline
\citep{geng2025l2m2}; transferable general multiagent benchmark & $\phi_{\mathrm{plan}}$ for high level subgoals above MARL execution & C2 planning; C5 learning; C6 coordination; C8 transfer & 1 & Return and success related performance in VMAS and MOSMAC; less than 20\% of baseline training samples in VMAS and tests with pretrained policies & Manufacturing task, inference latency distribution, hardware, safety layer and physical validation NR \\ \hline
\citep{lgpf2025swarm}; transferable swarm robotics benchmark & $\phi_{\mathrm{plan}}$ for language guided formation goals above MARL execution & C1 semantic grounding; C2 planning; C6 coordination; C8 transfer & 3 & Pattern formation quality and transfer from simulation to physical tabletop ``maru'' robots & Manufacturing task, parameter count, median and tail latency, deadline violations, safety assurance and continuous operation NR \\ \hline
\end{longtable}
}

\bibliographystyle{plainnat}
\bibliography{refs}

\end{document}